\documentclass{webofc}

\usepackage[varg]{txfonts}
\usepackage{natbib}
\usepackage{lineno}
\usepackage{hyperref}
\usepackage{amstext}
\usepackage{array}
\usepackage{hhline}
\usepackage{amsmath}
\usepackage{indentfirst}
\usepackage[symbol]{footmisc}

\newcolumntype{L}{>{$}l<{$}}

\newcommand{\hlsfml}{{\tt{hls4ml\,}}}
\newcommand{\pysr}{{\tt{PySR\,}}}

\newcommand{\qkeras}{{\tt{QKeras\,}}}

\newcommand{\github}{{\tt{github\,}}}

\begin{document}

\title{Symbolic Regression on FPGAs for Fast Machine Learning Inference}
\author{
\firstname{Ho Fung} \lastname{Tsoi}\inst{1}\thanks{\email{ho.fung.tsoi@cern.ch}} \and 
\firstname{Adrian Alan} \lastname{Pol}\inst{2} \and 
\firstname{Vladimir} \lastname{Loncar}\inst{3,4} \and 
\firstname{Ekaterina} \lastname{Govorkova}\inst{3}  
\and \firstname{Miles} \lastname{Cranmer}\inst{2,5} \and 
\firstname{Sridhara} \lastname{Dasu}\inst{1} \and 
\firstname{Peter} \lastname{Elmer}\inst{2}
\and \firstname{Philip} \lastname{Harris}\inst{3} 
\and \firstname{Isobel} \lastname{Ojalvo}\inst{2}\and 
\firstname{Maurizio} \lastname{Pierini}\inst{6} 
}
\institute{
University of Wisconsin-Madison, USA \and 
Princeton University, USA \and 
Massachusetts Institute of Technology, USA 
\and
Institute of Physics Belgrade, Serbia
\and
Flatiron Institute, USA
\and
European Organization for Nuclear Research (CERN), Switzerland}

\abstract{
The high-energy physics community is investigating the potential of deploying machine-learning-based solutions on Field-Programmable Gate Arrays (FPGAs) to enhance physics sensitivity while still meeting data processing time constraints.
In this contribution, we introduce a novel end-to-end procedure that utilizes a machine learning technique called symbolic regression (SR).
It searches the equation space to discover algebraic relations approximating a dataset.
We use \pysr (a software to uncover these expressions based on an evolutionary algorithm) and extend the functionality of \hlsfml (a package for machine learning inference in FPGAs) to support \pysr-generated expressions for resource-constrained production environments. Deep learning models often optimize the top metric by pinning the network size because the vast hyperparameter space prevents an extensive search for neural architecture.
Conversely, SR selects a set of models on the Pareto front, which allows for optimizing the performance-resource trade-off directly.
By embedding symbolic forms, our implementation can dramatically reduce the computational resources needed to perform critical tasks.
We validate our method on a physics benchmark: the multiclass classification of jets produced in simulated proton-proton collisions at the CERN Large Hadron Collider.
We show that our approach can approximate a 3-layer neural network using an inference model that achieves up to a 13-fold decrease in execution time, down to 5 ns, while still preserving more than 90\% approximation accuracy.
}
\maketitle


\section{Introduction}\label{SEC:Introduction}

Symbolic regression (SR) is a machine learning technique that seeks to discover mathematical expressions that best fit a dataset.
The outcome of SR is an analytic equation that captures the underlying patterns and relationships within the data. As the equations are interpretable, SR can provide valuable insights into natural sciences, including high-energy physics (HEP).
Furthermore, by allowing the selection of models on the Pareto front, SR enables the optimization of the performance-resource trade-off, making it a promising alternative to other machine learning methods, especially deep learning models.
This is a crucial feature in the context of the Large Hadron Collider (LHC) experiments, which must process proton-proton collisions at a 40 MHz rate and tens of terabytes of raw data per second.
This extreme data rate and the current size of the buffering system impose a maximum latency of $\mathcal{O}(1)$ $\mu$s for the real-time classification and filtering of data at the edge (or the {\em trigger system})~\cite{ATLAS:2020esi,Atlas:2137107,CMS:2020cmk,Zabi:2020gjd}.
In these conditions, lightweight algorithms running on custom hardware such as Field-Programmable Gate Arrays (FPGAs) for ultra low-latency inference are desired.

In this paper, we extend the functionality of the \hlsfml\footnote{\url{https://github.com/fastmachinelearning/hls4ml}}~\cite{vloncar_2021_5680908,Duarte:2018ite} (High Level Synthesis for Machine Learning) framework to provide parsing capabilities for the equation chosen by SR and High Level Synthesis (HLS) support for mathematical functions.
Our implementation is validated on a physics benchmark, demonstrating the effectiveness and potential of this approach to address the challenges faced by the HEP community.
For generating the expressions, we have chosen to utilize \pysr\footnote{\url{https://github.com/MilesCranmer/PySR}}~\cite{Cranmer:pysr}, an open-source software tool for SR that employs an evolutionary algorithm.
\pysr offers a comprehensive implementation of SR and is built on {\tt Julia} but interfaced from {\tt Python}, making it easily accessible and usable for practitioners in a wide range of fields, including HEP.
An example code is available on \github\footnote{\url{https://github.com/fastmachinelearning/hls4ml-tutorial}}.
The remainder of this paper is structured as follows.
Section~\ref{SEC:Dataset} introduces the dataset and the baseline model.
Section~\ref{SEC:Experiments} presents our implementations and results.
Lastly, Section~\ref{SEC:Summary} summarizes the work and suggests future directions.

\section{Benchmark and Baseline}\label{SEC:Dataset}
To demonstrate the application of SR, we choose the jet identification problem from the HEP field.
A jet refers to a narrow cone of outgoing particles, and the process of identifying the original particle that initiated this collimated shower of particles with adjacent trajectories is called {\em jet tagging}.
Jets are central to many physics data analyses at the LHC experiments.
The data for this case study are generated from simulated jets that result from the decay and hadronization of quarks and gluons produced in high-energy collisions at the LHC.
The task is to tag a given jet as originating from either a quark ($q$), gluon ($g$), W boson ($W$), Z boson ($Z$), or top quark ($t$).
The dataset is publicly accessible from Zenodo~\cite{Pierini:LHCjet}.
A variety of jet recombination algorithms and substructure tools are implemented to build a list of 16 physics-motivated expert features: $\big($$\sum z\,\text{log}\,z$, $C_{1}^{\beta=0,1,2}$, $C_{2}^{\beta=1,2}$, $D_{2}^{\beta=1,2}$, $D_{2}^{(\alpha,\beta)=(1,1),(1,2)}$, $M_{2}^{\beta=1,2}$, $N_{2}^{\beta=1,2}$, $m_{\text{mMDT}}$, $\text{Multiplicity}$$\big)$, where the description of each of these variables is presented in Ref.~\cite{Coleman:2017fiq}.
The anti-$k_{\text{T}}$ algorithm~\cite{Cacciari:2008gp} with a distance parameter of $R = 0.8$ is used to cluster all jets. 
A cut on the reconstructed jet $p_{\text{T}}$ is applied to remove extreme events from the analysis~\cite{Duarte:2018ite}.
More detailed descriptions of the dataset can be found in Refs.~\cite{Duarte:2018ite,Moreno:2019bmu,Coleman:2017fiq}.

The architecture of the baseline model is adopted from Ref.~\cite{Duarte:2018ite}, which is a fully connected Neural Network (NN) consisting of three hidden layers of 64, 32, and 32 nodes, respectively, and ReLU activation functions.
The input layer takes the 16 high-level features as input, and the output layer consists of five nodes with a softmax activation function, yielding the probability of a jet originating from each of the five classes. 
This architecture was chosen to provide reasonable performance (75\% overall accuracy and 90\% per class accuracy) while keeping the model lightweight~\cite{Coelho:2020zfu,Duarte:2018ite,Loncar:2020hqp,Hawks:2021ruw}.
The model is trained with \qkeras\footnote{\url{https://github.com/google/qkeras}}, where the kernel weights, biases, and activation functions are quantized to a fixed precision and constrained during weight optimization, called {\em quantization-aware training} (QAT)~\cite{Coelho:2020zfu}.
This is necessary since post-training quantization (no fine-tuning) results in reduced accuracy. The baseline models presented in Section~\ref{SEC:Experiments} are fine-tuned for each precision considered.
For evaluation, the model is converted to HLS firmware using \hlsfml.


\section{Implementations and Results}\label{SEC:Experiments}
\begin{figure}[h!]
  \begin{center}
    \includegraphics[width=0.4\textwidth]{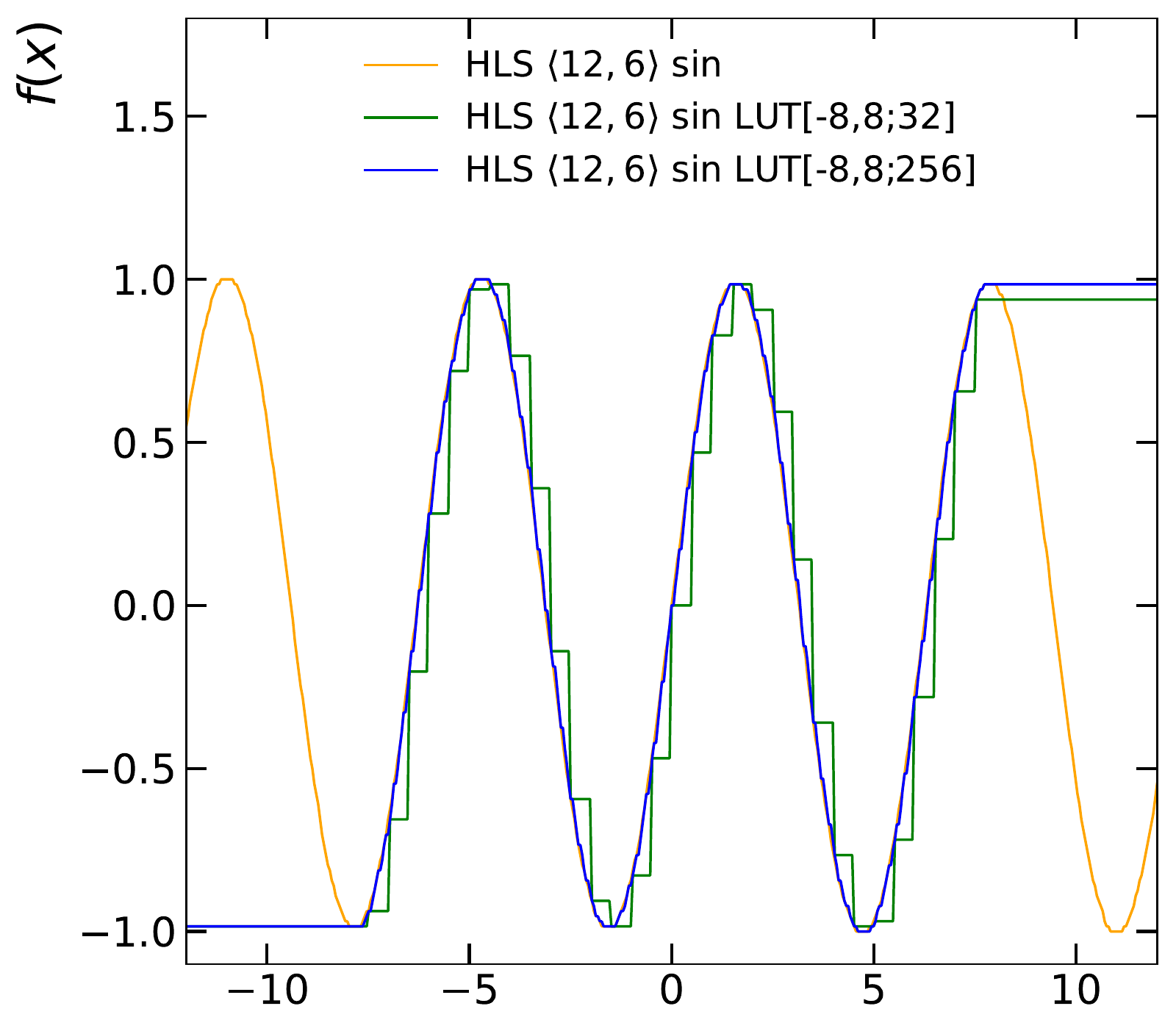}
    \includegraphics[width=0.4\textwidth]{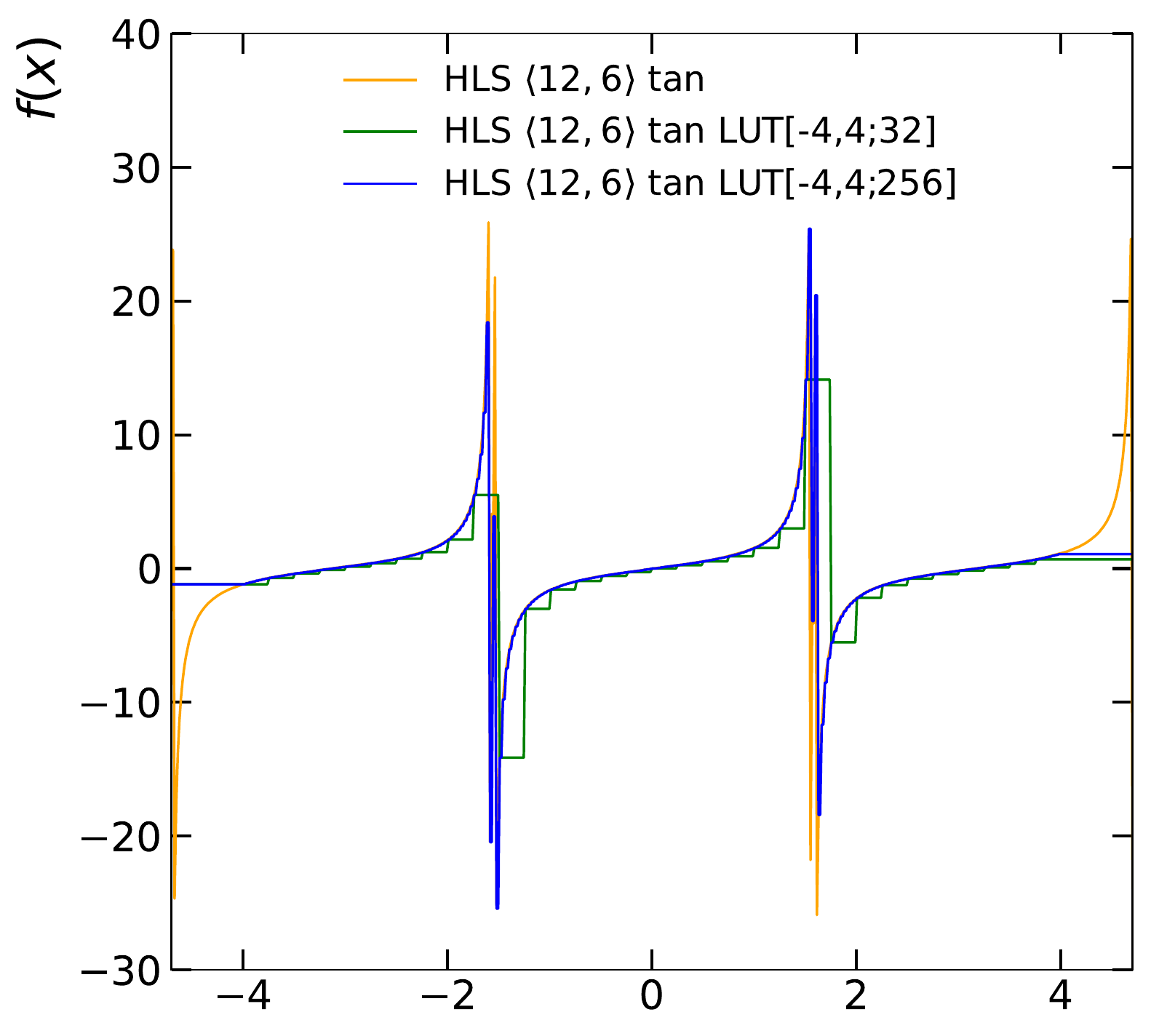}\\
    \includegraphics[width=0.4\textwidth]{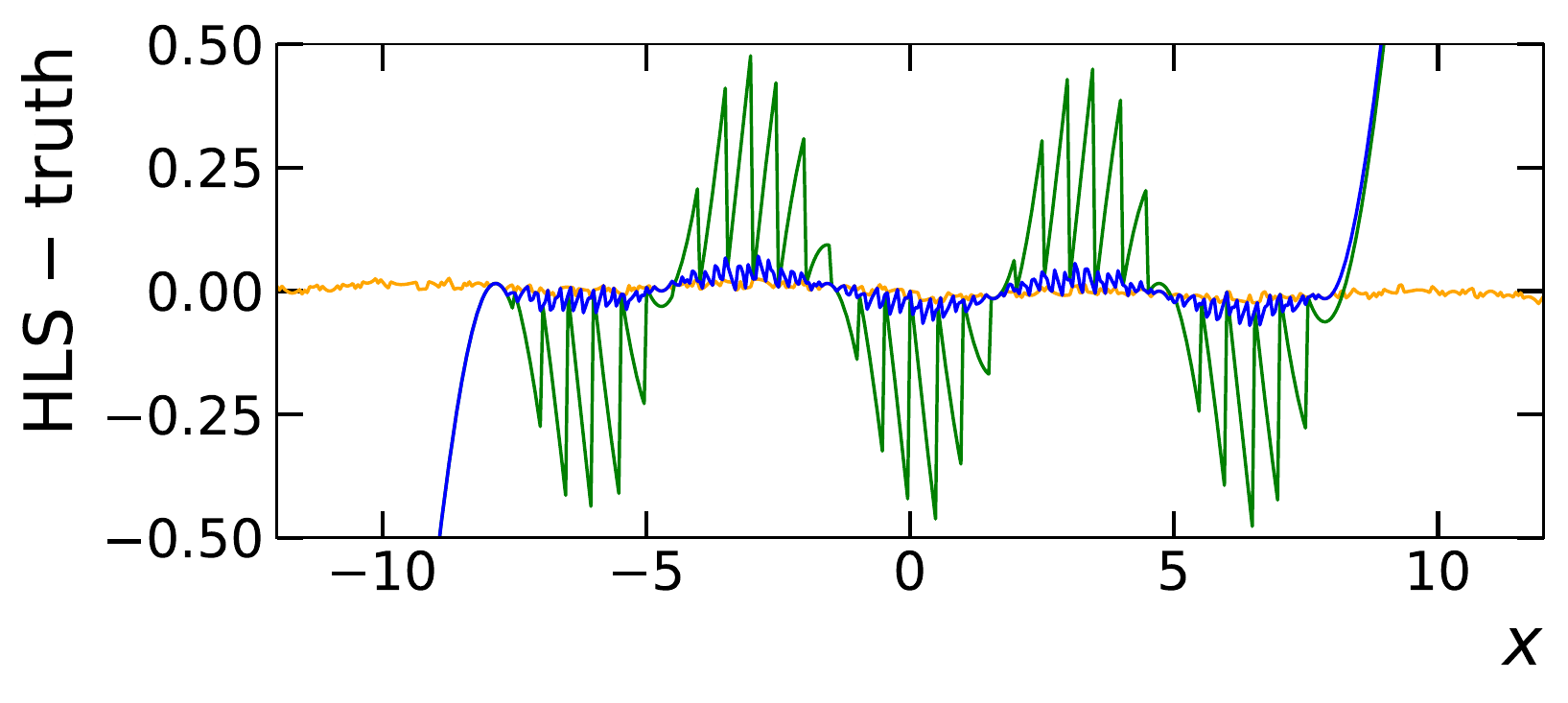}
    \includegraphics[width=0.4\textwidth]{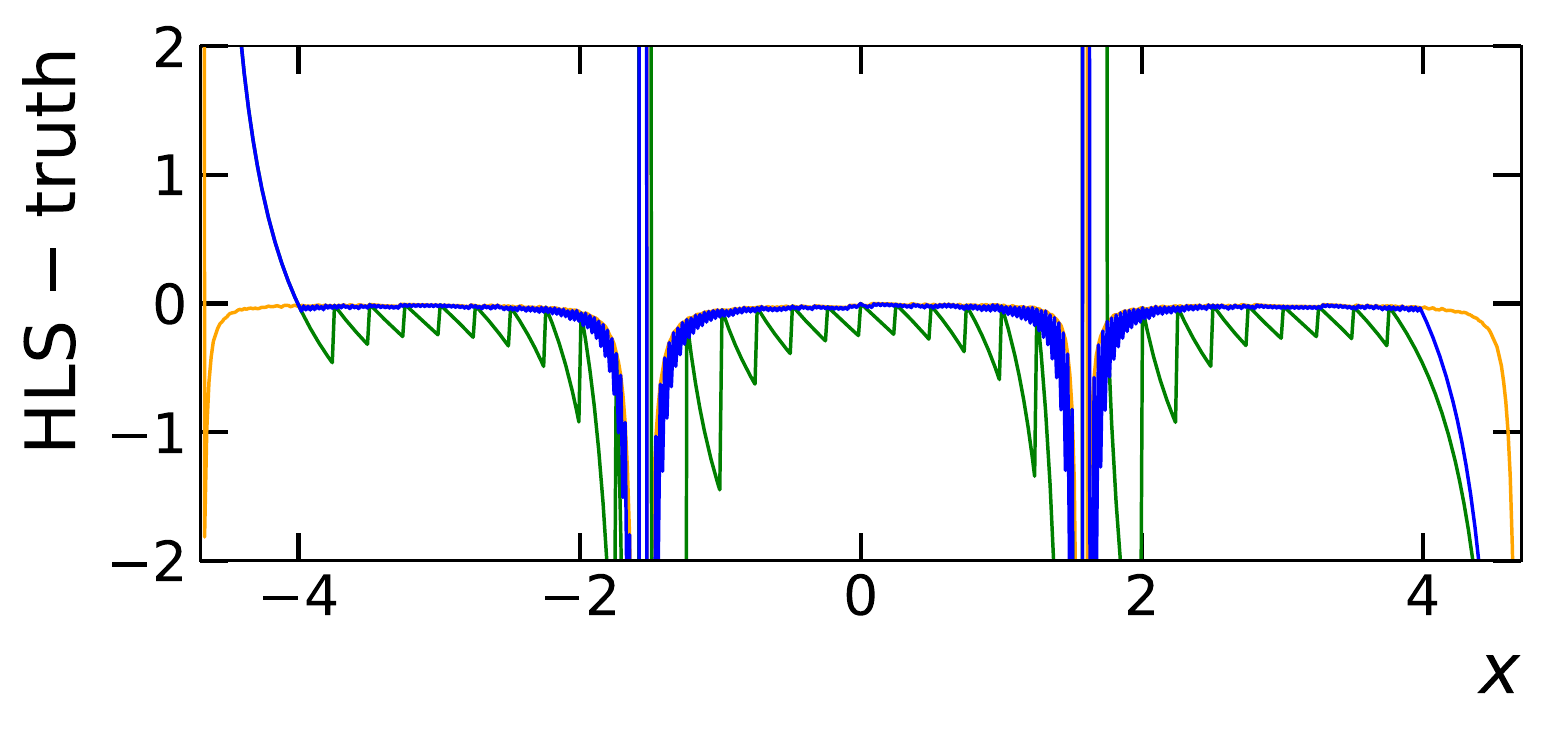}
    \caption{The sine (left) and tangent (right) functions evaluated with and without the use of LUTs, implemented in HLS with precision $\langle\text{12},\text{6}\rangle$, i.e., 12 bits variable with 6 integer bits. The LUT notation reads: $[$range start, range end; table size$]$ for table definition. The lower panel shows the function deviation from the truth.}
    \label{fig:LUTfunc}
  \end{center}
\end{figure}

To deploy symbolic expressions on FPGAs, we use the \hlsfml library.
We extended \hlsfml with support for expressions through the Xilinx HLS math library.
To further optimize resource utilization and reduce latency, we added functionality to enable approximation of mathematical functions with lookup tables (LUTs).
The comparison of LUT-based functions with HLS math library functions is illustrated in Fig.~\ref{fig:LUTfunc}.
We use $\langle\text{B},\text{I}\rangle$ to denote fixed point precision, where B is the total number of bits allocated or bit width, and I is the number of integer bits.

In the following experiments, we apply SR to fit the LHC jet dataset and demonstrate its resource efficiency in the context of FPGA deployment.
We consider models of five independent algebraic expressions as functions of the 16 high-level input features, $\hat{\mathbf{y}}=s(\mathbf{x})$ with $s:\mathbb{R}^{16}\rightarrow\mathbb{R}^{5}$, where the inputs are standardized and the outputs $\hat{\mathbf{y}}$ correspond to the score for one of the five jet classes.
A jet is identified as the class whose tagger yields the highest score.

The search for expressions is performed using the \pysr package.
It uses an evolutionary algorithm to construct symbolic expressions, by growing the tree structure using combinations of constants, variables, and operators ({\tt +}, {\tt -}, {\tt $\times$}, {\tt /}, {\tt $(\cdot)^2$}, {\tt $\text{sin}(\cdot)$}, etc.).
The search starts from a random combination without requiring {\it a priori} knowledge of the underlying functional form.
Expressions are evaluated by a specified metric, and the best ones are passed on to the next generation, where mutation (selecting one node to modify) and crossover (swapping the subtrees of two solutions) can take place to explore more combinations.
The \pysr measure of complexity, denoted as $c$, is set to 1 by default for each constant, variable, and operator included in an expression.
The complexity of an expression is the sum of the complexity of all its components.
We set the model selection strategy so that the candidate model with the lowest loss will be selected regardless of complexity, as long as it does not exceed the maximum value, $c_{\text{max}}$, of our choice.
In this setting, the algorithm attempts to solve the following optimization problem for the dataset $\{(\mathbf{x}^{i},\mathbf{y}^{i})\}$ with each input $\mathbf{x}^{i}\in\mathbb{R}^{16}$ and label $\mathbf{y}^{i}\in\mathbb{R}^{5}$:
\begin{equation}
  h^{*}_{g},h^{*}_{q},h^{*}_{t},h^{*}_{W},h^{*}_{Z}=\operatorname*{arg\,min}_{h_{g},h_{q},h_{t},h_{W},h_{Z}\in\mathcal{S}_{c_{\text{max}}}} \sum_{i}\sum_{f\in\{g,q,t,W,Z\}}\ell(h_{f}(\mathbf{x}^{i}),y^{i}_{f}),
\end{equation}
where $\mathcal{S}_{c_{\text{max}}}$ is the space of equations (i.e., $h_{f}:\mathbb{R}^{16}\rightarrow\mathbb{R}$) with complexity ranging from 1 to $c_{\text{max}}$ satisfying all constraints specified in the configuration (choice of operators, function nesting, etc.).
We use the L2 margin loss, $\ell$ given by
\begin{equation}
  \ell(\hat{y}_{f},y_{f})=(1-\hat{y}_{f}y_{f})^2,\,\text{with label\ }y_{f}=
  \begin{cases}
    +1,&\text{if }f\text{ corresponds to the true jet class}\\
    -1,&\text{otherwise}
  \end{cases}.
\end{equation}
The selection of this loss is due to its domain being $\mathbb{R}^2$, since the model outputs are not restricted to any fixed range.

The downside of using evolutionary algorithms for SR is that it is a complex combinatorial problem which does not scale well to high-dimensional datasets.
To alleviate this challenge, \pysr uses a random forest regressor to evaluate the relative importance of each input feature.
We ask \pysr to select 6 out of the 16 inputs available for model training in the following experiments.

For resource estimation, each model is converted to FPGA firmware using \hlsfml, which is then synthesized with Vivado HLS (2020.1)~\cite{vivado}, targeting a Xilinx Virtex UltraScale+ VU9P FPGA with part number `xcvu9p-flga2577-2-e'.
All results are derived after the HLS compilation step.
The clock frequency is set to 200 MHz (or clock period of 5 ns), which is typical for the LHC real-time trigger environment~\cite{ATLAS:2020esi,Atlas:2137107,CMS:2020cmk,Zabi:2020gjd}.
The initiation interval is set to 1.
In the following studies, we monitor the accuracy, latency, and resource usage (digital signal processors, or DSPs, and LUTs) to compare the models.

\subsection{Plain implementation}\label{SEC:plain}

We first study models with a single class of mathematical functions: polynomial, trigonometric, exponential, and logarithmic.
For the polynomial model, only arithmetic operators are considered: {\tt +}, {\tt -}, and {\tt $\times$}.
For other models, an additional operator is added, respectively: {\tt$\text{sin}(\cdot)$} for trigonometric, {\tt$\text{Gauss}(\cdot)=\text{exp}(-(\cdot)^{2})$} for exponential, and {\tt$\text{log}(\text{abs}(\cdot))$} for logarithmic.
For simplicity, function nesting (e.g., {\tt$\text{sin}(\text{sin}(\cdot))$}) is not allowed.
Each operator has a complexity of 1 by default.
Searches are repeated for $c_{\text{max}}=$ 20, 40, and 80, to observe how model accuracy and resource usage change with model size.
Table~\ref{tab:expr} shows the expressions per class for the trigonometric model with $c_{\text{max}}=20$.
Table~\ref{tab:expr_topquark} shows expressions for the $t$ tagger in all models with $c_{\text{max}}=40$.
Accuracy is shown in Fig.~\ref{fig:plainandLUT_acc}. FPGA resource usage and latency are shown in Fig.~\ref{fig:plainandLUT_resource}.
\begin{table}[h!]
  \centering
  \resizebox{0.75\textwidth}{!}{
    \begin{tabular}{ L | L | L }
      \hhline{===}
      \text{Tagger} & \text{Expression for the trigonometric model with $c_{\text{max}}=20$} & \text{AUC} \\ \hhline{===}
      g & \text{sin}(- 2C_{1}^{\beta=1} + 0.31C_{1}^{\beta=2} + m_{\text{mMDT}} + \text{Multiplicity} - 0.09\text{Multiplicity}^2 - 0.79) & 0.897 \\ \hline
      q & -0.33(\text{sin}(m_{\text{mMDT}}) - 1.54)(\text{sin}(- C_{1}^{\beta=1} + C_{1}^{\beta=2} + \text{Multiplicity}) - 0.81)\text{sin}(m_{\text{mMDT}}) - 0.81 & 0.853 \\ \hline
      t & \text{sin}(C_{1}^{\beta=1} + C_{1}^{\beta=2} - m_{\text{mMDT}} + 0.22(C_{1}^{\beta=2} - 0.29)(- C_{1}^{\beta=2} + C_{2}^{\beta=1} - \text{Multiplicity}) - 0.68) & 0.920 \\ \hline
      W & -0.31(\text{Multiplicity} + (2.09 - \text{Multiplicity})\text{sin}(8.02C_{1}^{\beta=2} + 0.98)) - 0.5 & 0.877 \\ \hline
      Z & (\text{sin}(4.84m_{\text{mMDT}}) + 0.59)\text{sin}(m_{\text{mMDT}} + 1.14)\text{sin}(C_{1}^{\beta=2} + 4.84m_{\text{mMDT}}) - 0.94 & 0.866 \\ \hline
      \hhline{===}
    \end{tabular}
  }
  \caption{Expressions generated by \pysr for the trigonometric model with $c_{\text{max}}=20$. The operator complexity is set to 1 by default. Constants are rounded to two decimal places for readability. The area under the receiver operating characteristic (ROC) curve, or AUC, is reported.}
  \label{tab:expr}
\end{table}

\begin{table}[h!]
  \centering
  \resizebox{\textwidth}{!}{
    \begin{tabular}{ L | L | L }
      \hhline{===}
      \text{Model} & \text{Expression for the $t$ tagger with $c_{\text{max}}=40$} & \text{AUC} \\ \hhline{===}
      \text{Polynomial} & C_{1}^{\beta=2} + 0.09m_{\text{mMDT}}(2C_{1}^{\beta=1} + M_{2}^{\beta=2} - m_{\text{mMDT}} - \text{Multiplicity} - (1.82C_{1}^{\beta=1} - M_{2}^{\beta=2})(C_{1}^{\beta=2} - 0.49m_{\text{mMDT}}) - 3.22) - 0.53 & 0.914 \\ \hline
      \text{Trigonometric} & \text{sin}(0.06(\sum z\,\text{log}\,z)M_{2}^{\beta=2} - 0.25C_{1}^{\beta=2}(- C_{1}^{\beta=1} + 2C_{1}^{\beta=2} - M_{2}^{\beta=2} + \text{Multiplicity} - 8.86) - m_{\text{mMDT}} + 0.06\text{Multiplicity} - 0.4) & 0.925 \\ \hline
      \text{Exponential} & 0.23C_{1}^{\beta=1}(-m_{\text{mMDT}} + \text{Gauss}(0.63\text{Multiplicity}) + 1) - \text{Gauss}(C_{1}^{\beta=1}) + 0.45C_{1}^{\beta=2} - 0.23m_{\text{mMDT}} & 0.920 \\ & +\, 0.23\text{Gauss}((4.24 - 1.19C_{2}^{\beta=1})(C_{1}^{\beta=2} - m_{\text{mMDT}})) + 0.15 & \\ \hline
      \text{Logarithmic} & C_{1}^{\beta=2} - 0.1m_{\text{mMDT}}(\text{Multiplicity}\times\text{log}(\text{abs}(\text{Multiplicity})) + 2.2) - 0.02\text{log}(\text{abs}(\text{Multiplicity})) & 0.923 \\ & -\, 0.1(C_{1}^{\beta=2}(C_{1}^{\beta=1} - 1.6M_{2}^{\beta=2} + m_{\text{mMDT}} + 1.28) - m_{\text{mMDT}} - 0.48)\text{log}(\text{abs}(C_{1}^{\beta=2})) - 0.42 & \\
      \hhline{===}
    \end{tabular}
  }
  \caption{Expressions generated by \pysr for the $t$ tagger in different models with $c_{\text{max}}=40$. Operator complexity is set to 1 by default. Constants are rounded to two decimal places for readability.}
  \label{tab:expr_topquark}
\end{table}

\subsection{Function approximation with LUTs}
\begin{figure}[!h]
  \begin{center}
    \includegraphics[width=0.4\textwidth]{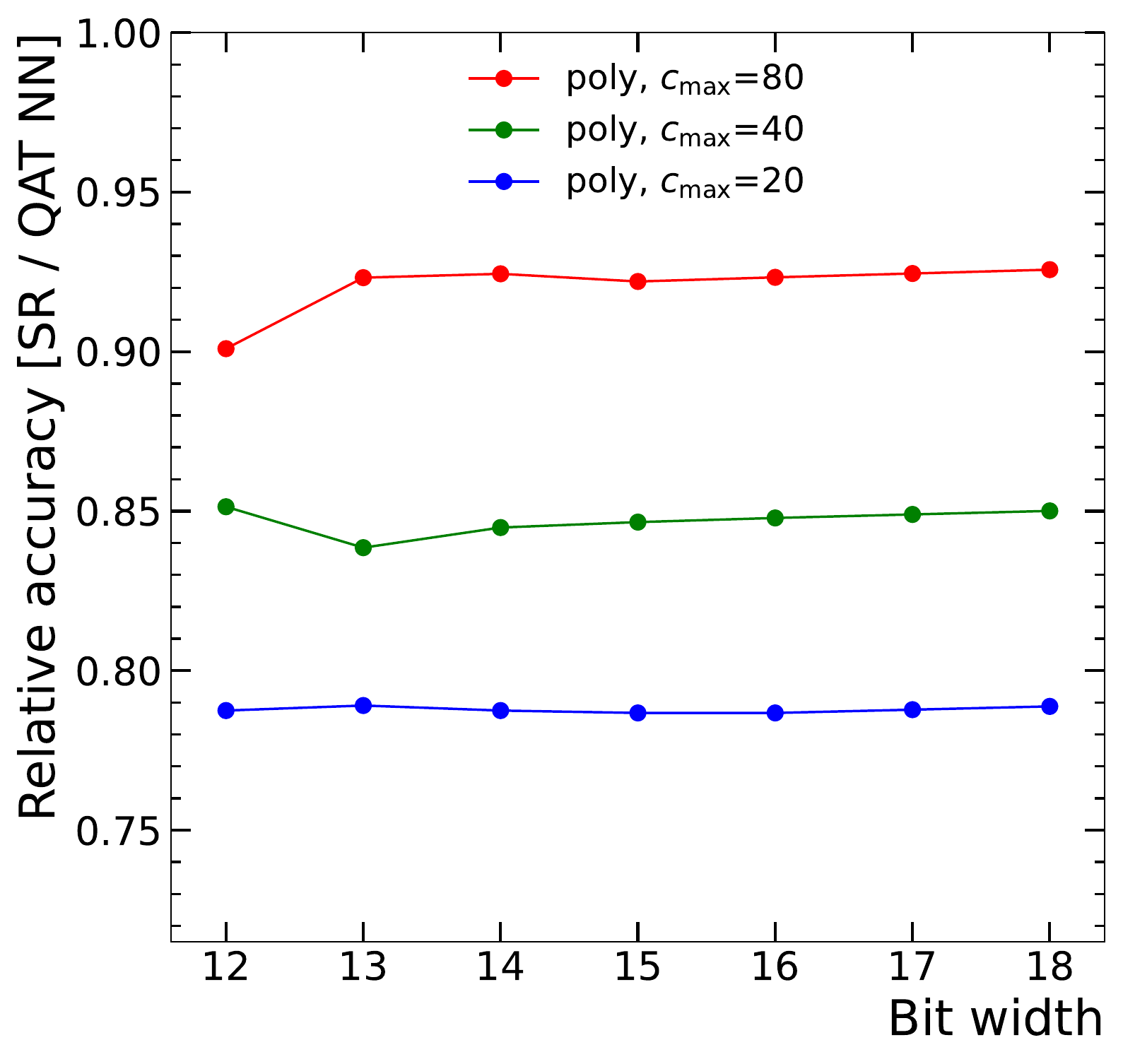}
    \includegraphics[width=0.4\textwidth]{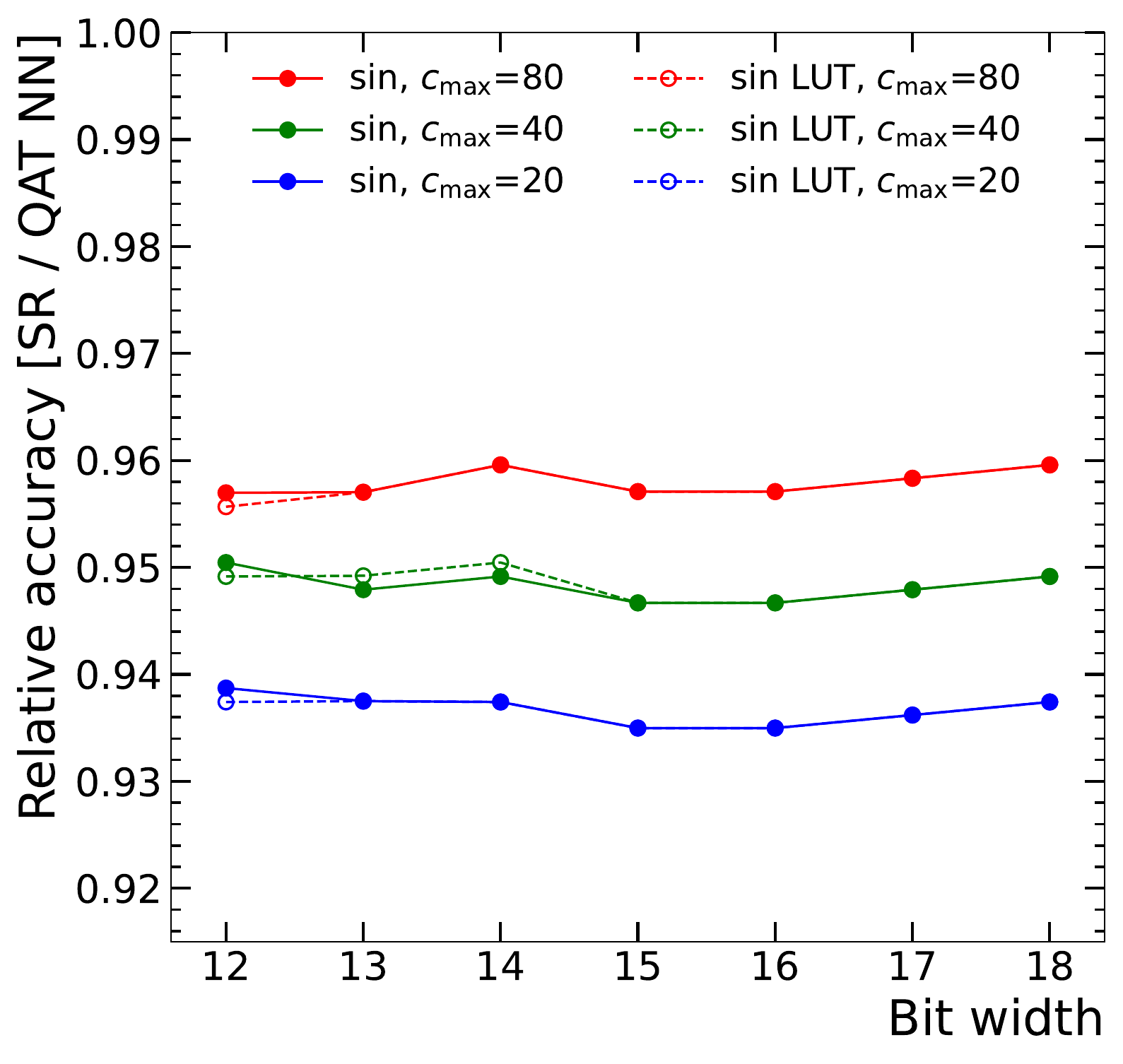}\\
    \includegraphics[width=0.4\textwidth]{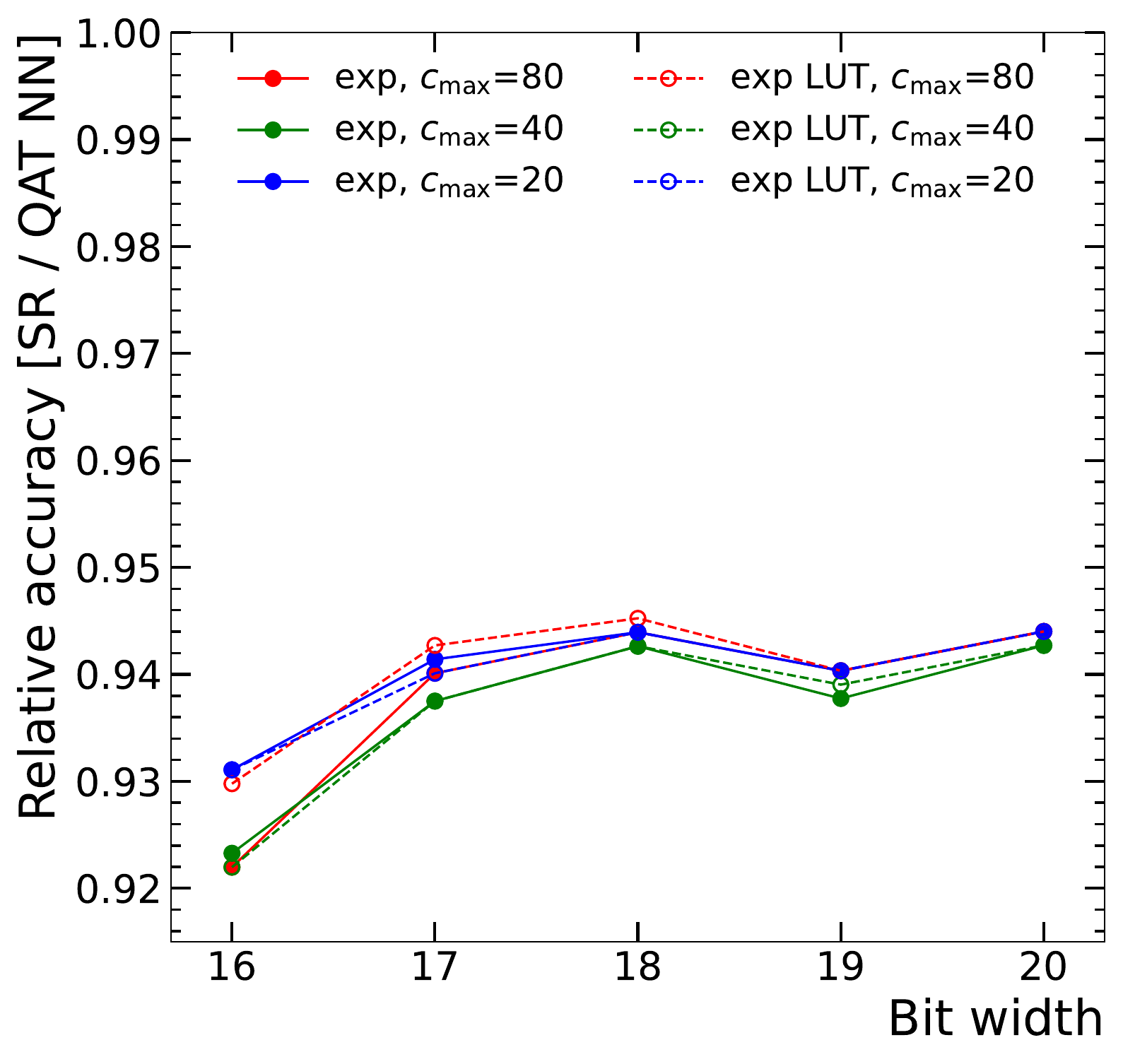}
    \includegraphics[width=0.4\textwidth]{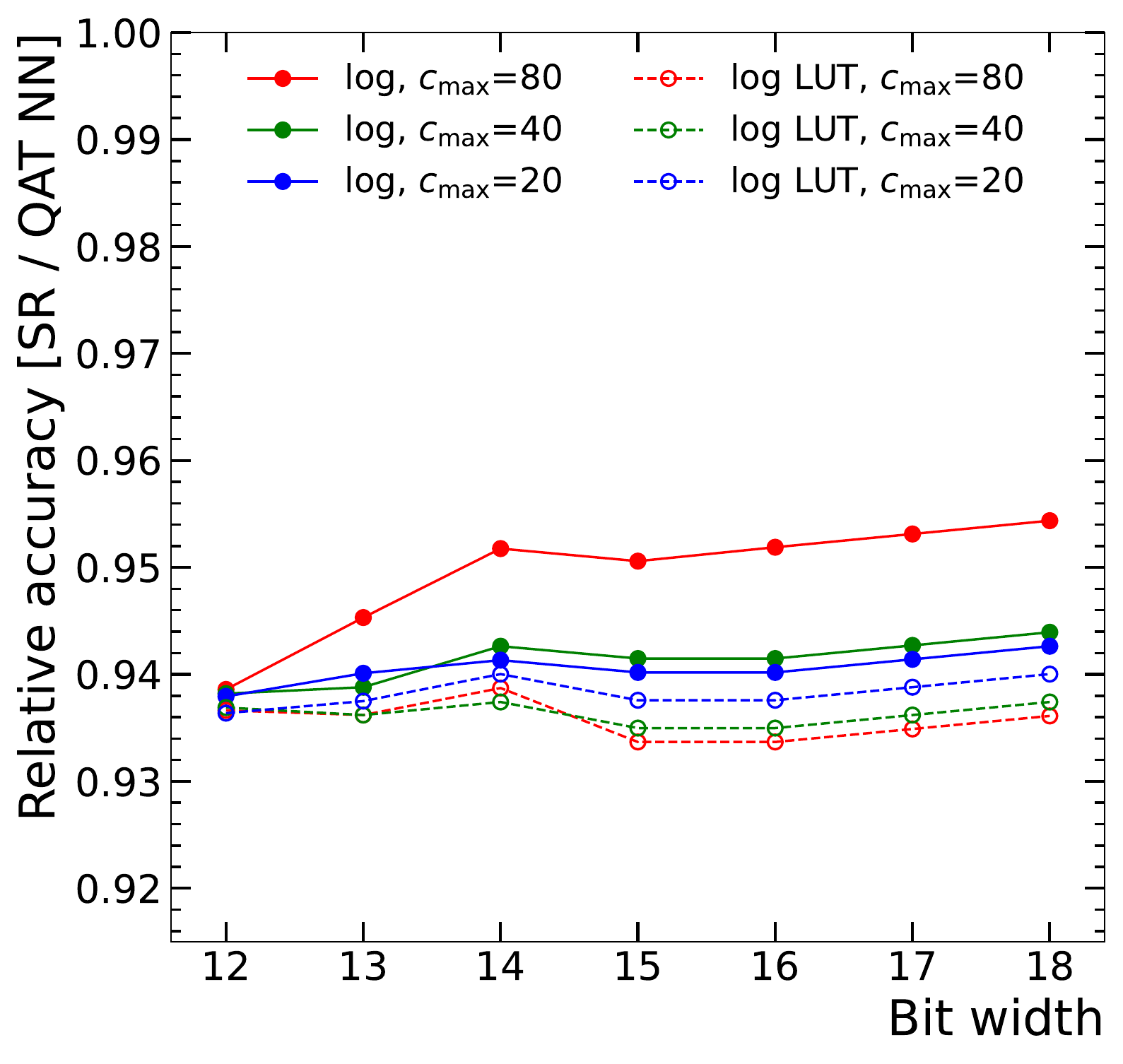}
    \caption{Relative accuracy as a function of bit width, for polynomial (top left), trigonometric (top right), exponential (bottom left), and logarithmic (bottom right) models. The relative accuracy is evaluated with respect to the baseline QAT NN trained and implemented at corresponding precision. The number of integer bits is fixed at $I=12$ for the exponential model and at $I=6$ for other models.}
    \label{fig:plainandLUT_acc}
  \end{center}
\end{figure}

\begin{figure}[!h]
  \begin{center}
    \includegraphics[width=0.329\textwidth]{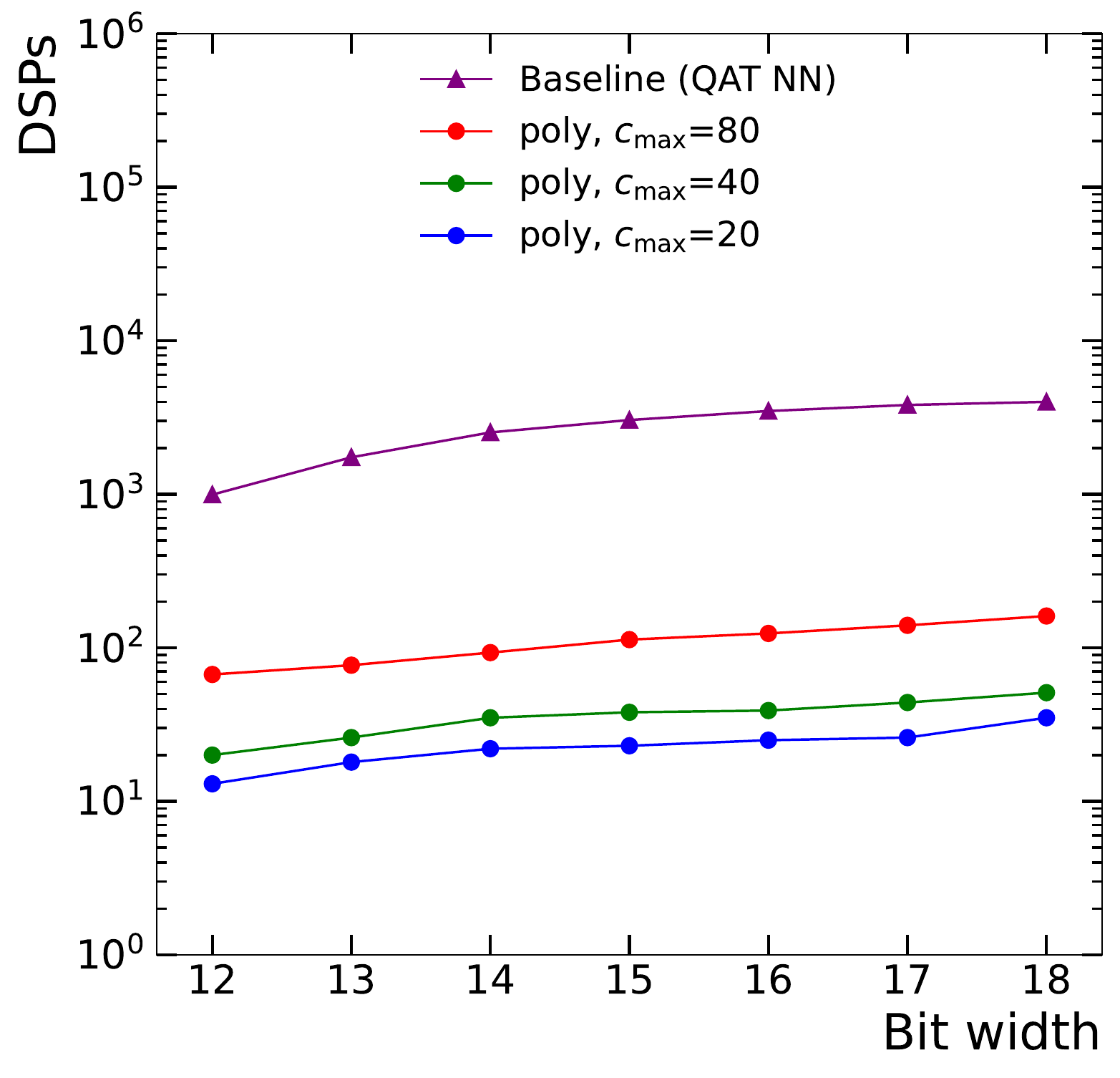}
    \includegraphics[width=0.329\textwidth]{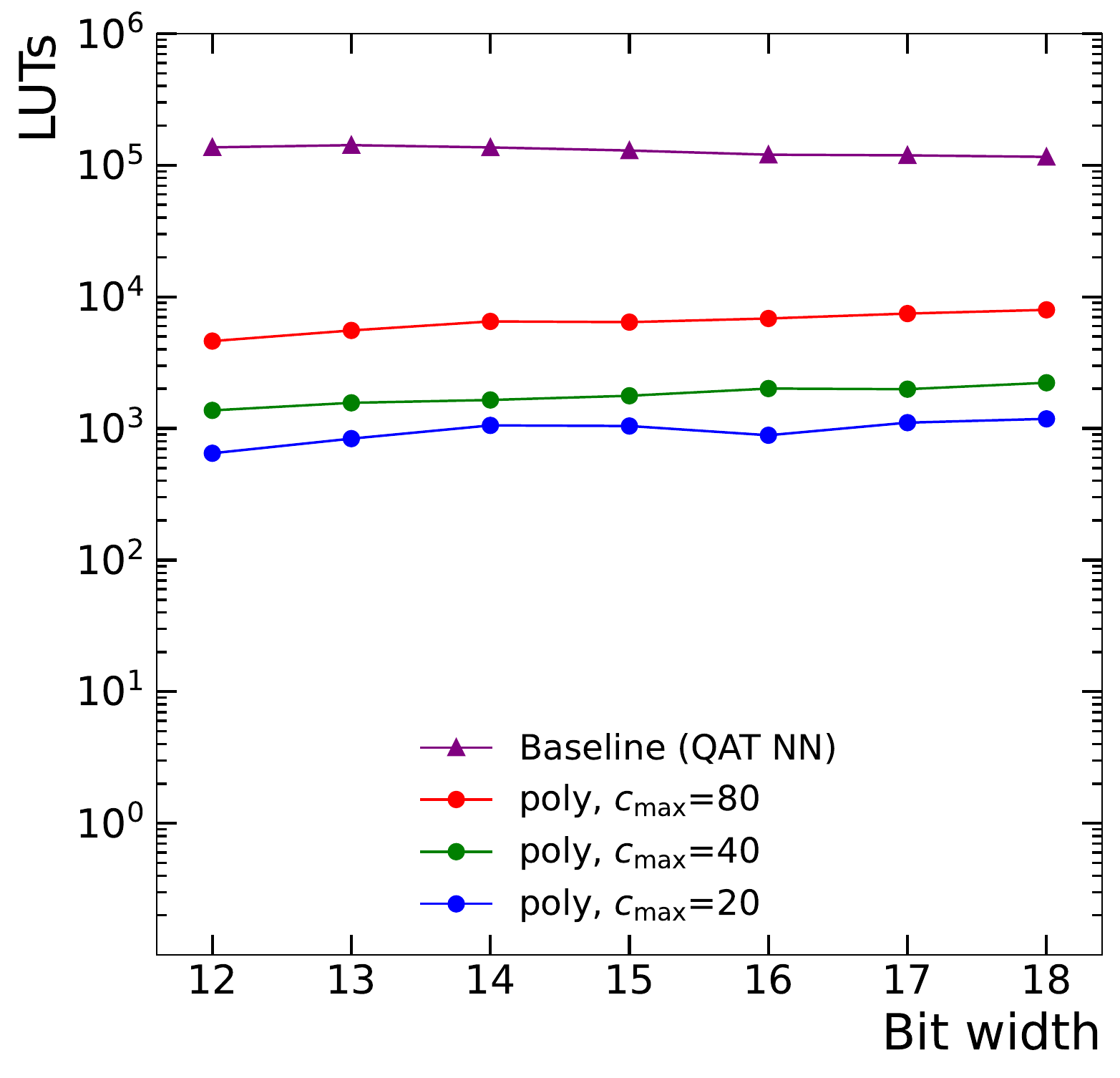}
    \includegraphics[width=0.329\textwidth]{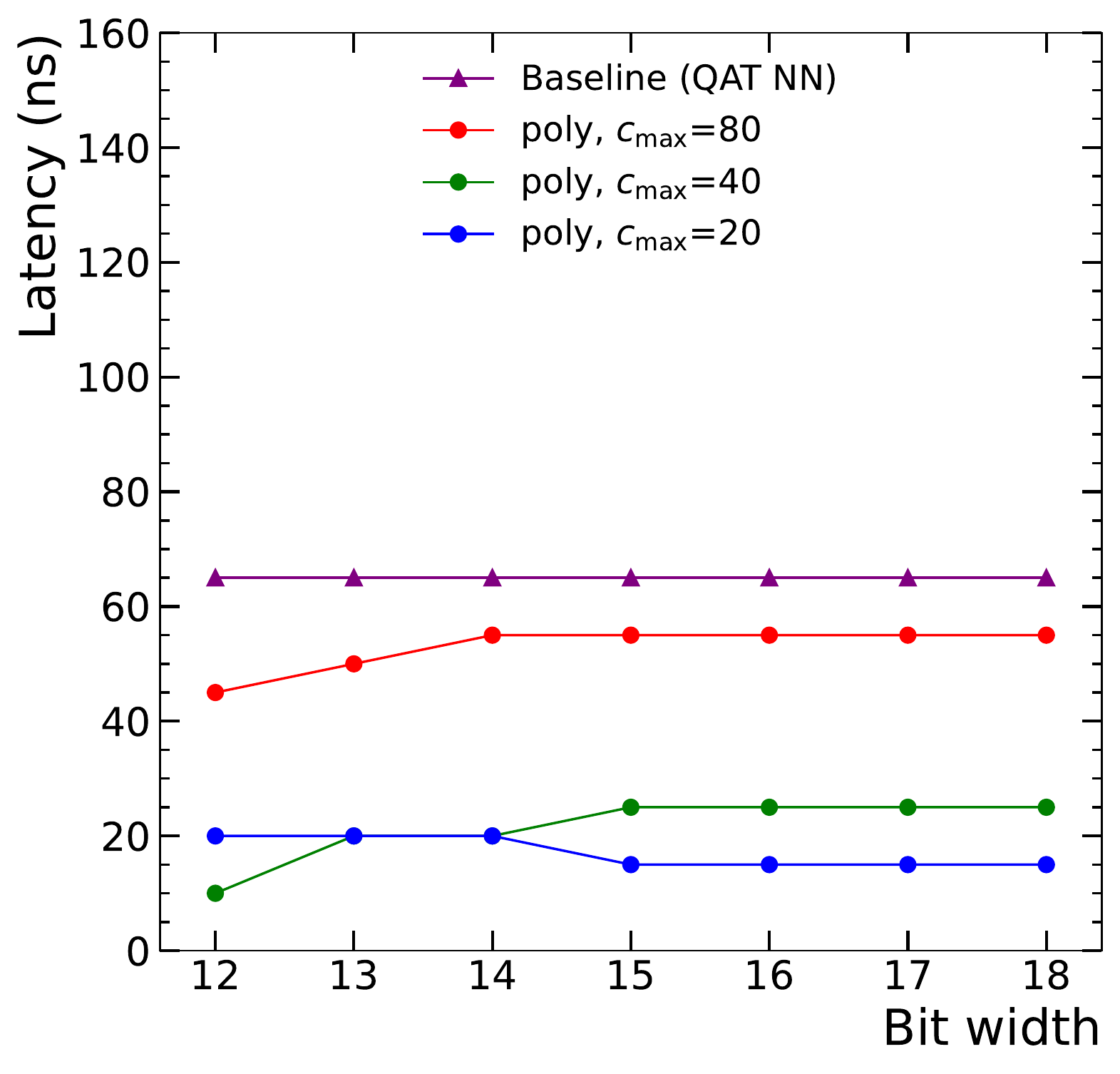}\\
    \includegraphics[width=0.329\textwidth]{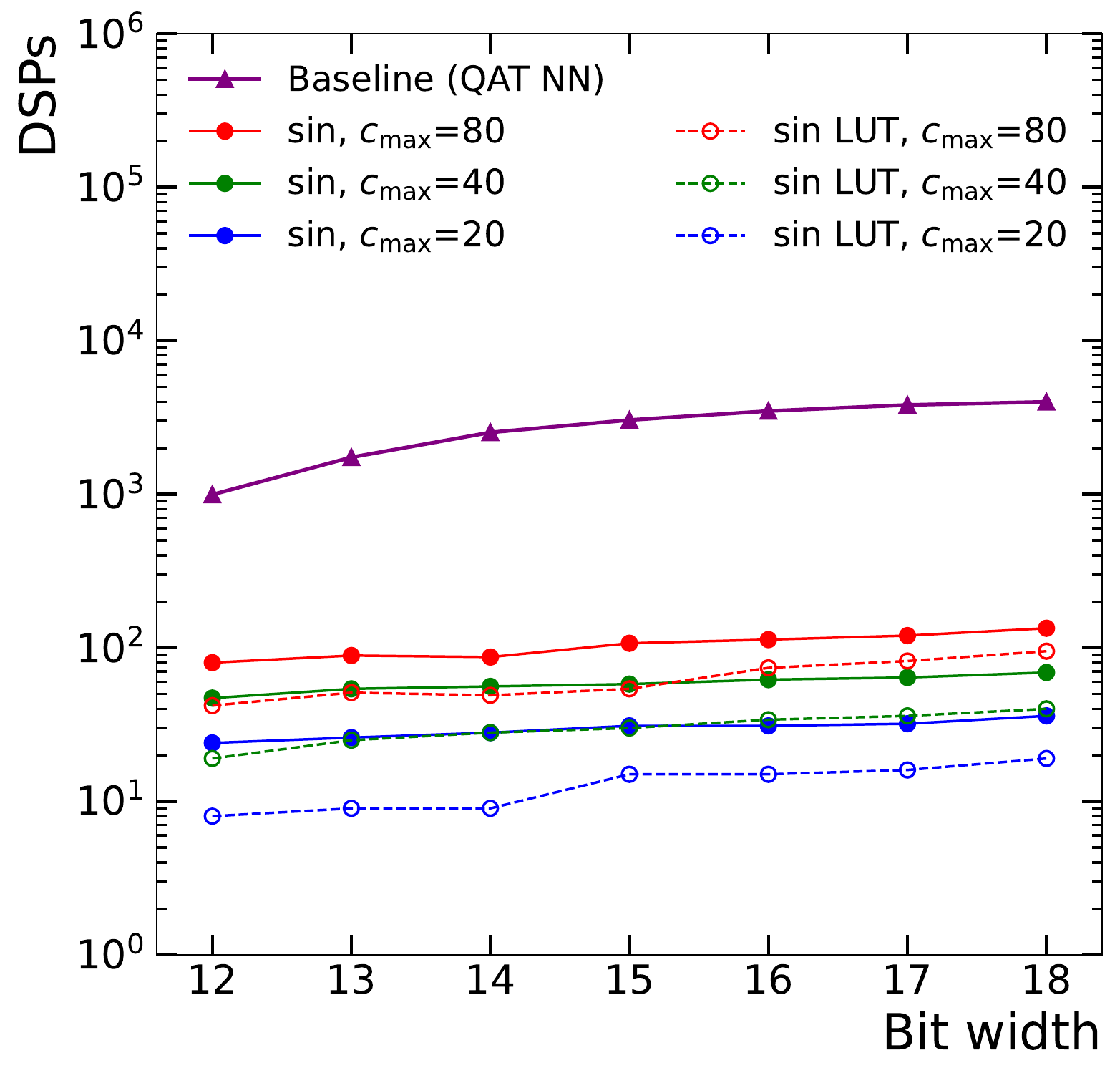}
    \includegraphics[width=0.329\textwidth]{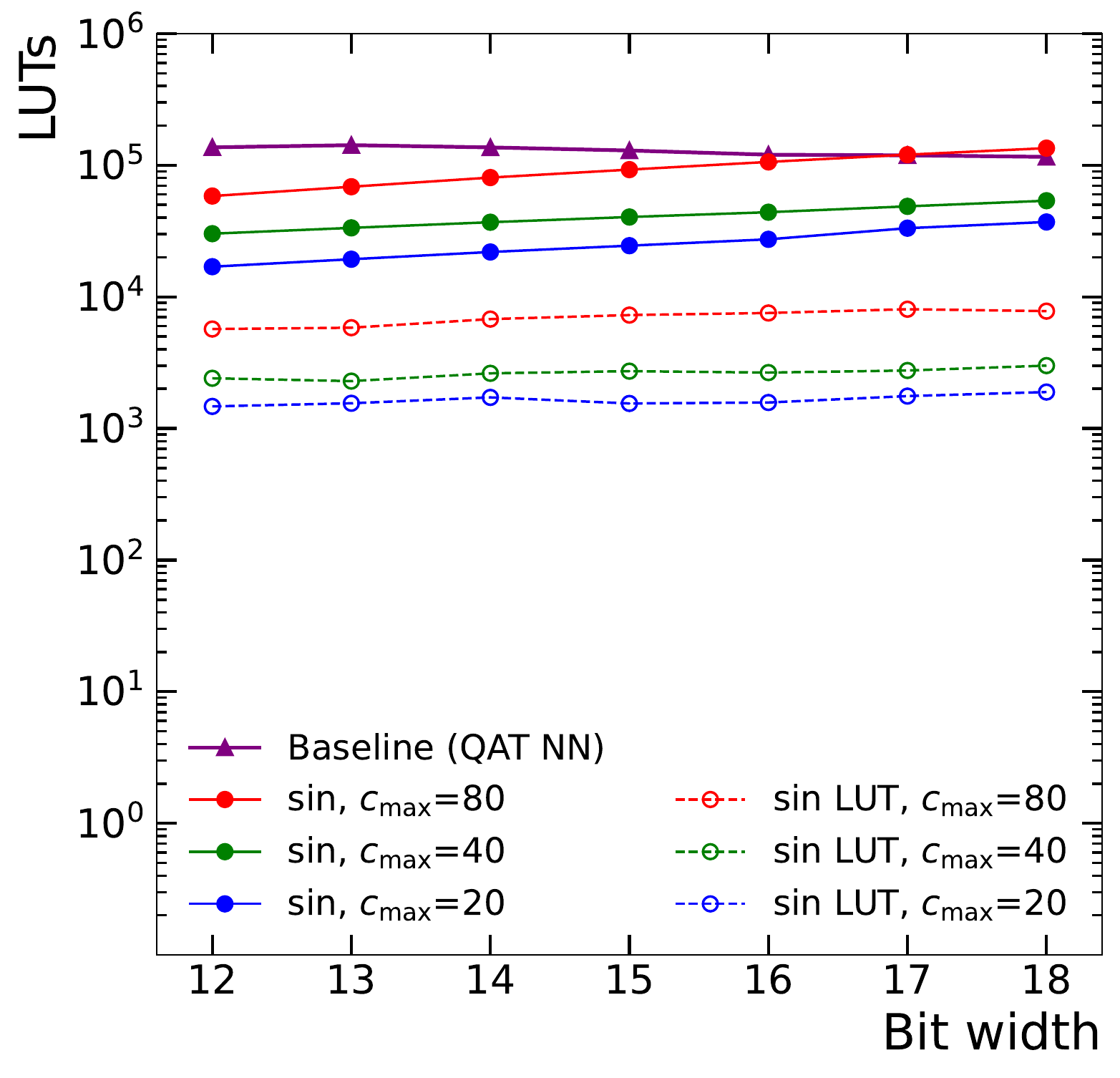}
    \includegraphics[width=0.329\textwidth]{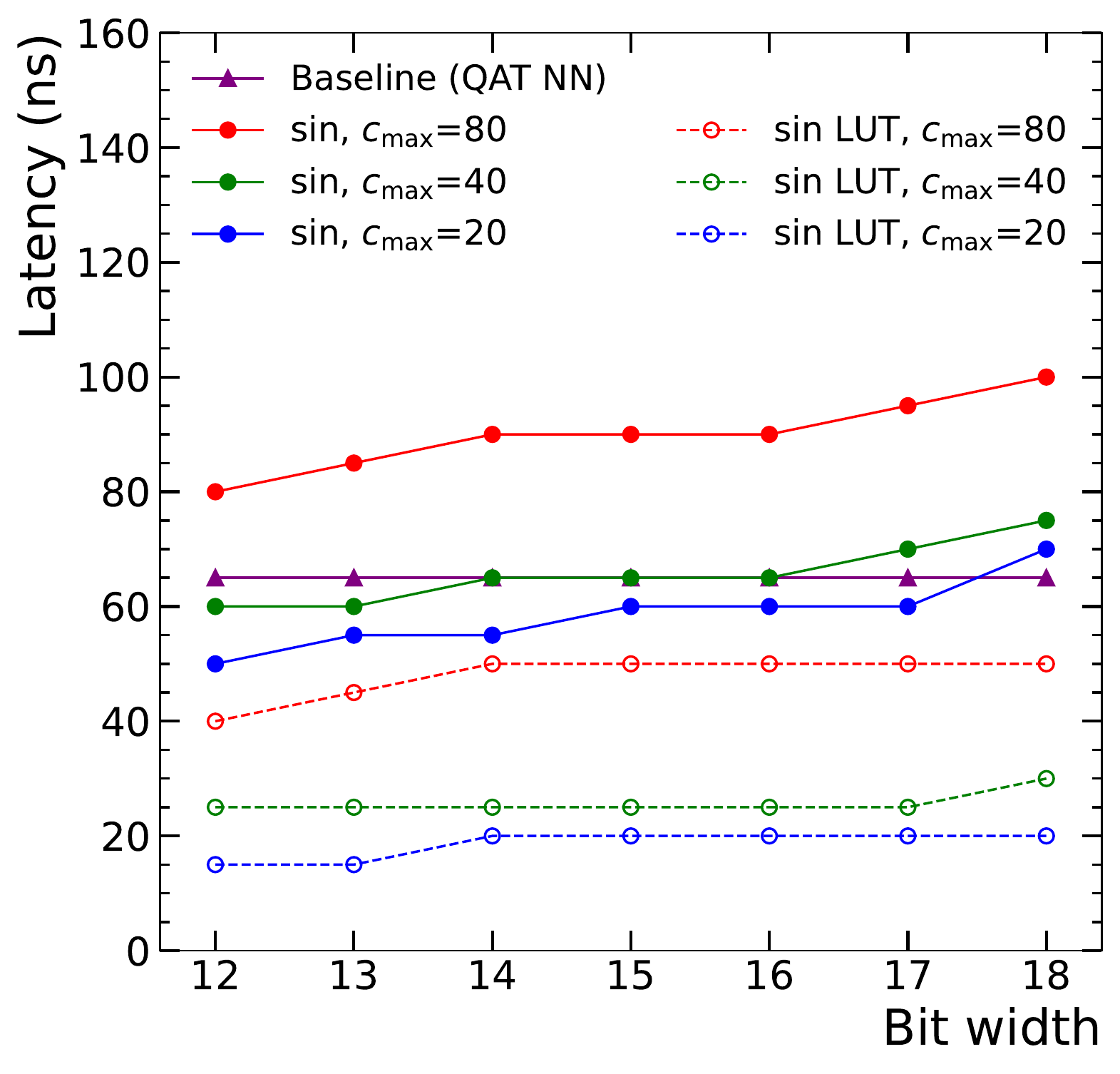}\\
    \includegraphics[width=0.329\textwidth]{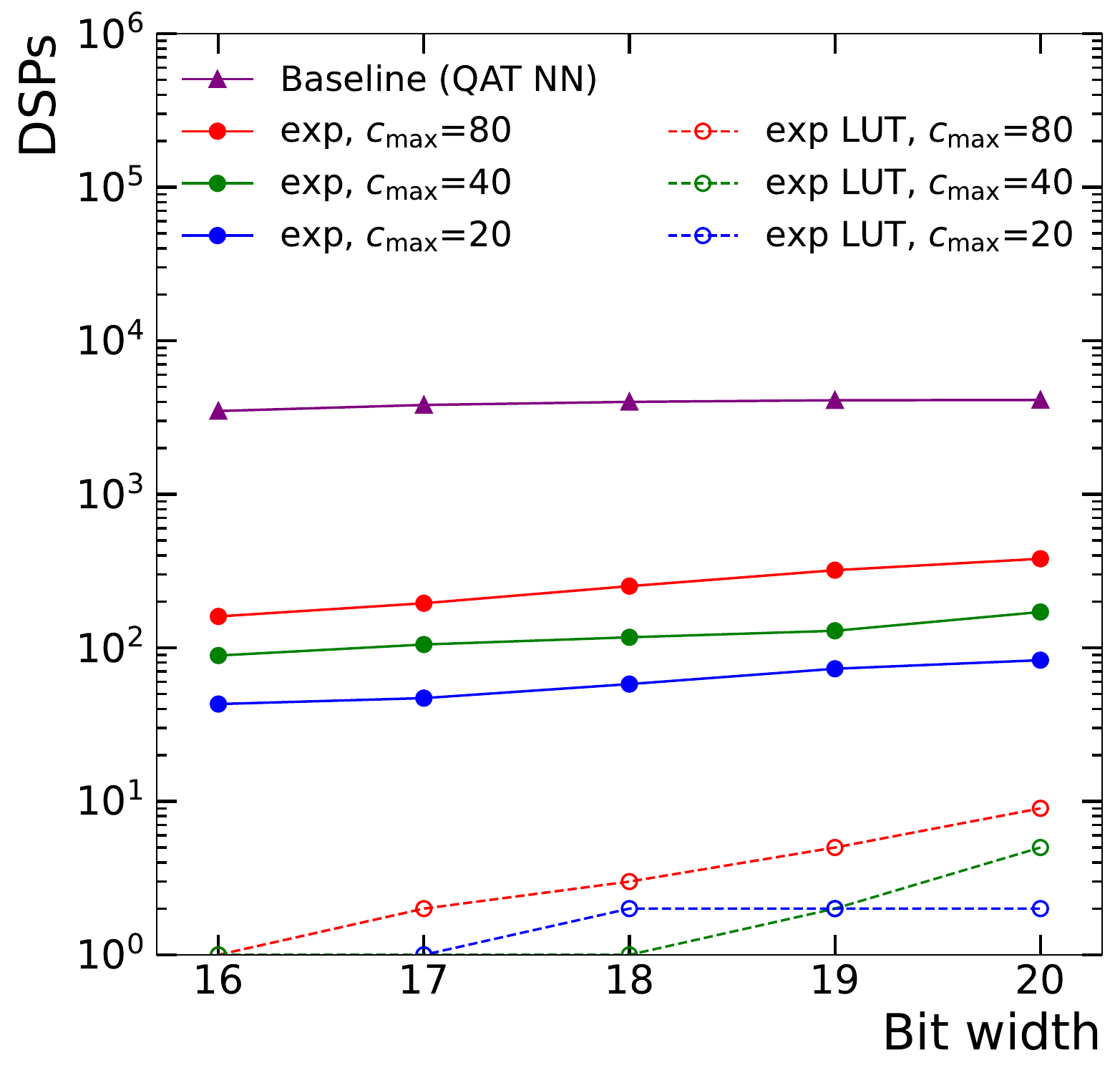}
    \includegraphics[width=0.329\textwidth]{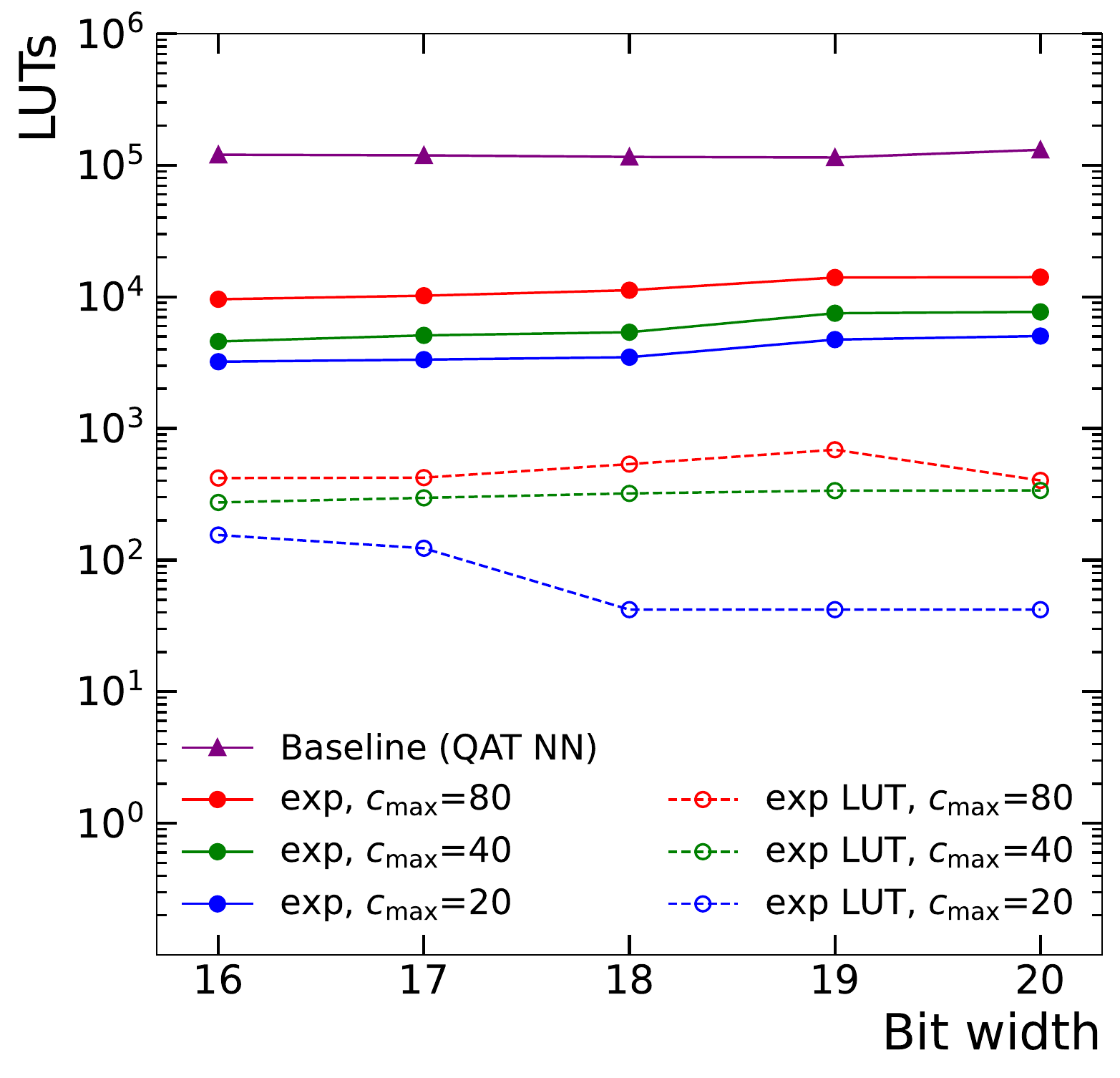}
    \includegraphics[width=0.329\textwidth]{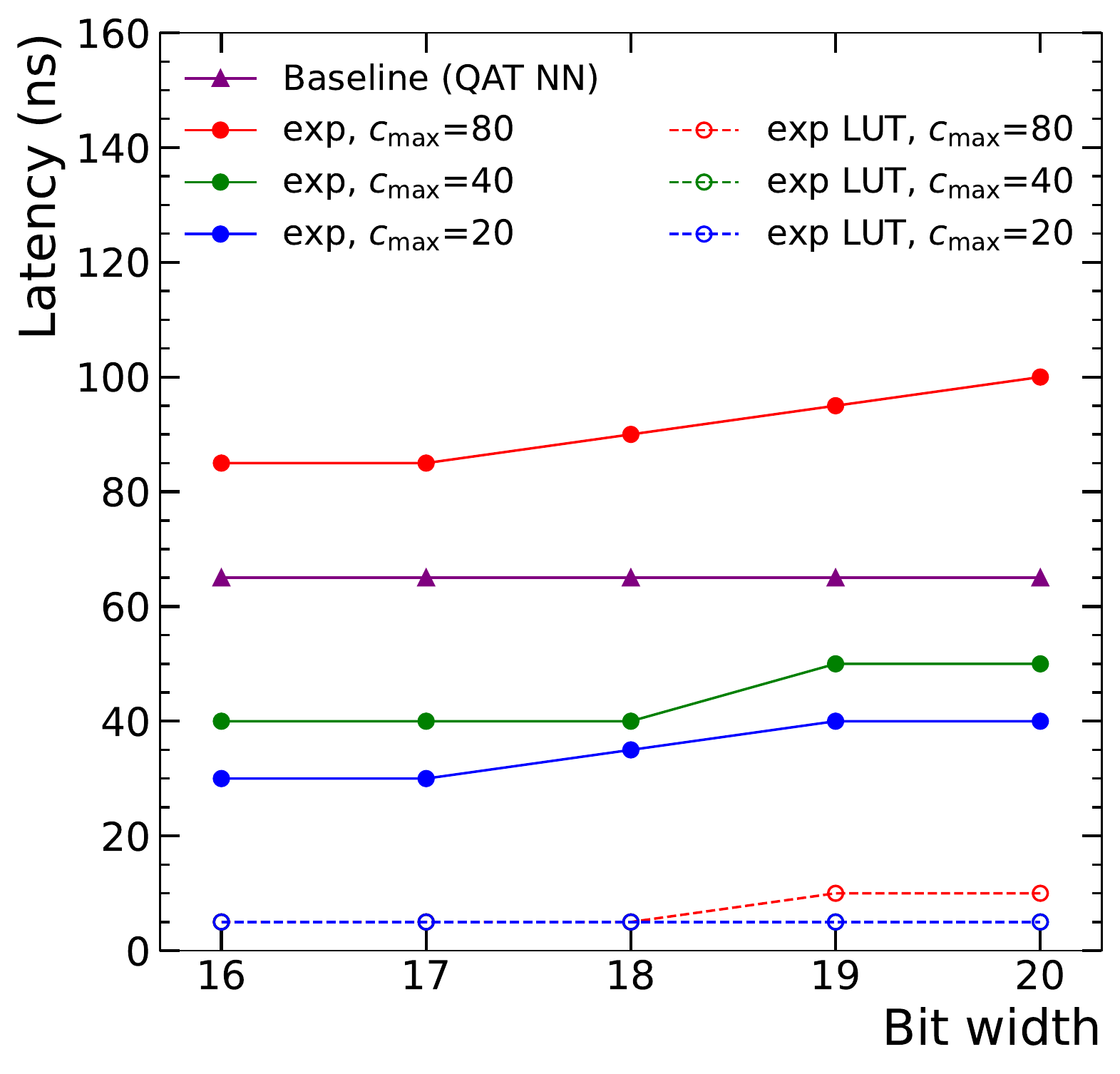}\\
    \includegraphics[width=0.329\textwidth]{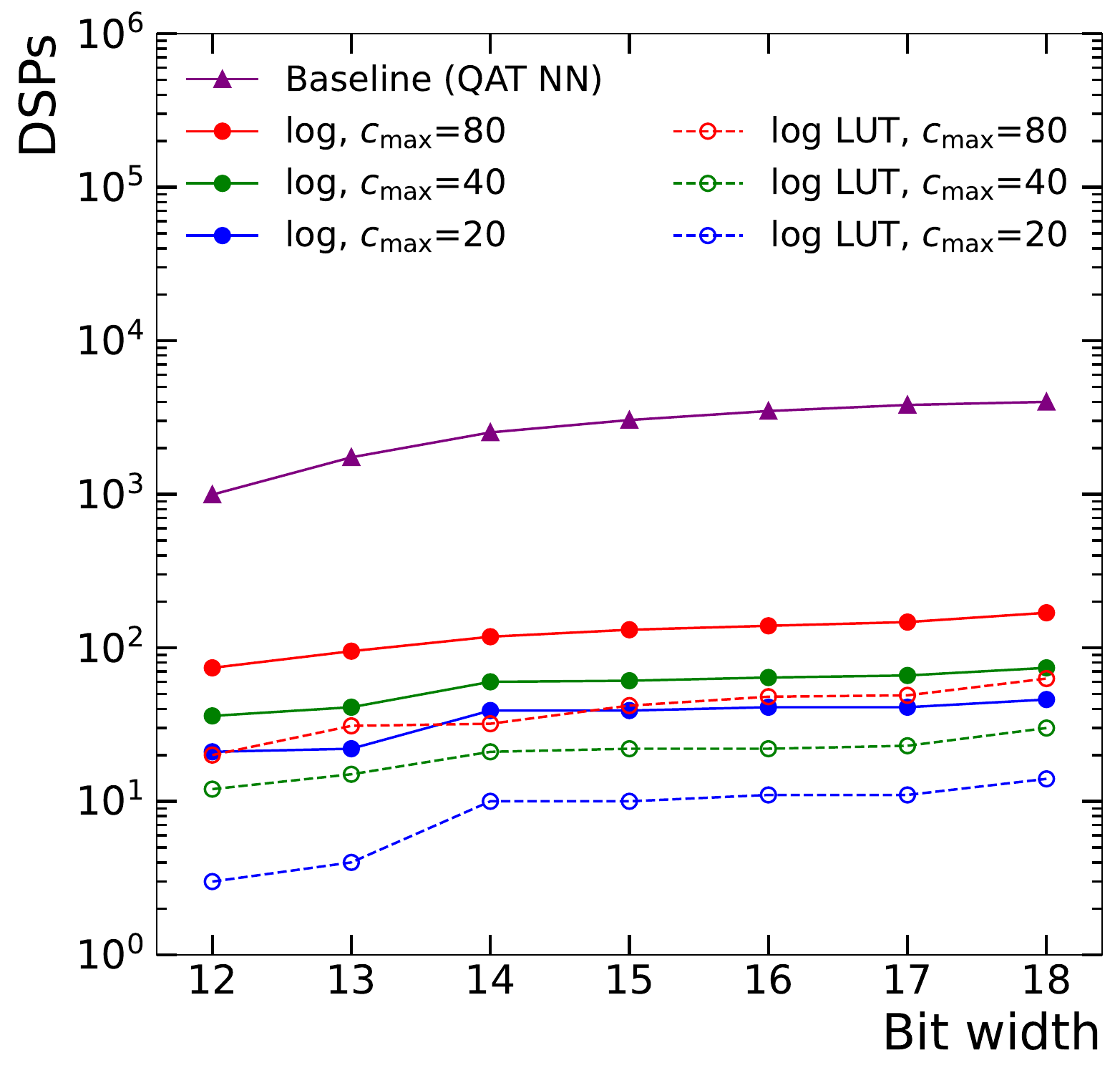}
    \includegraphics[width=0.329\textwidth]{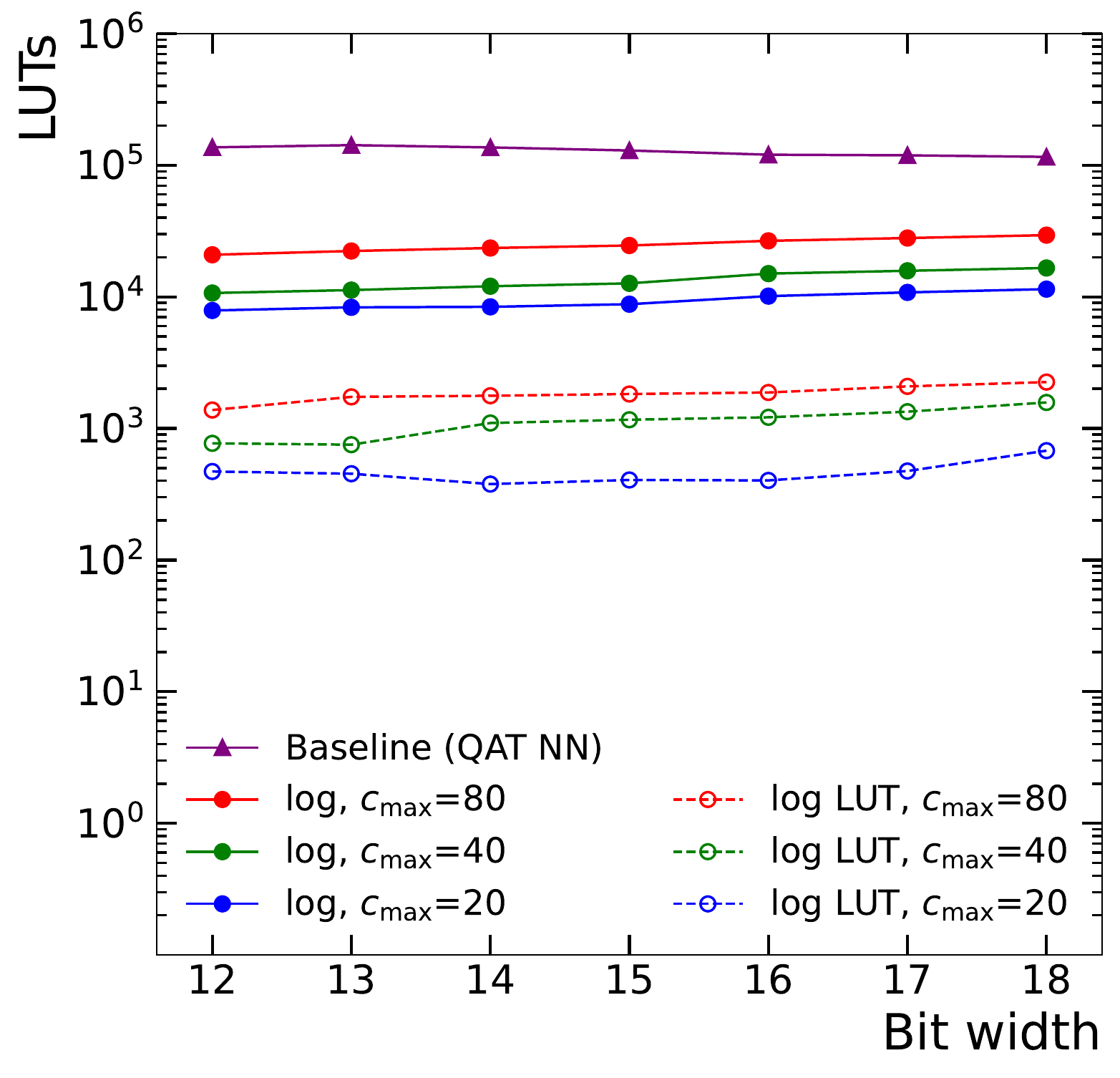}
    \includegraphics[width=0.329\textwidth]{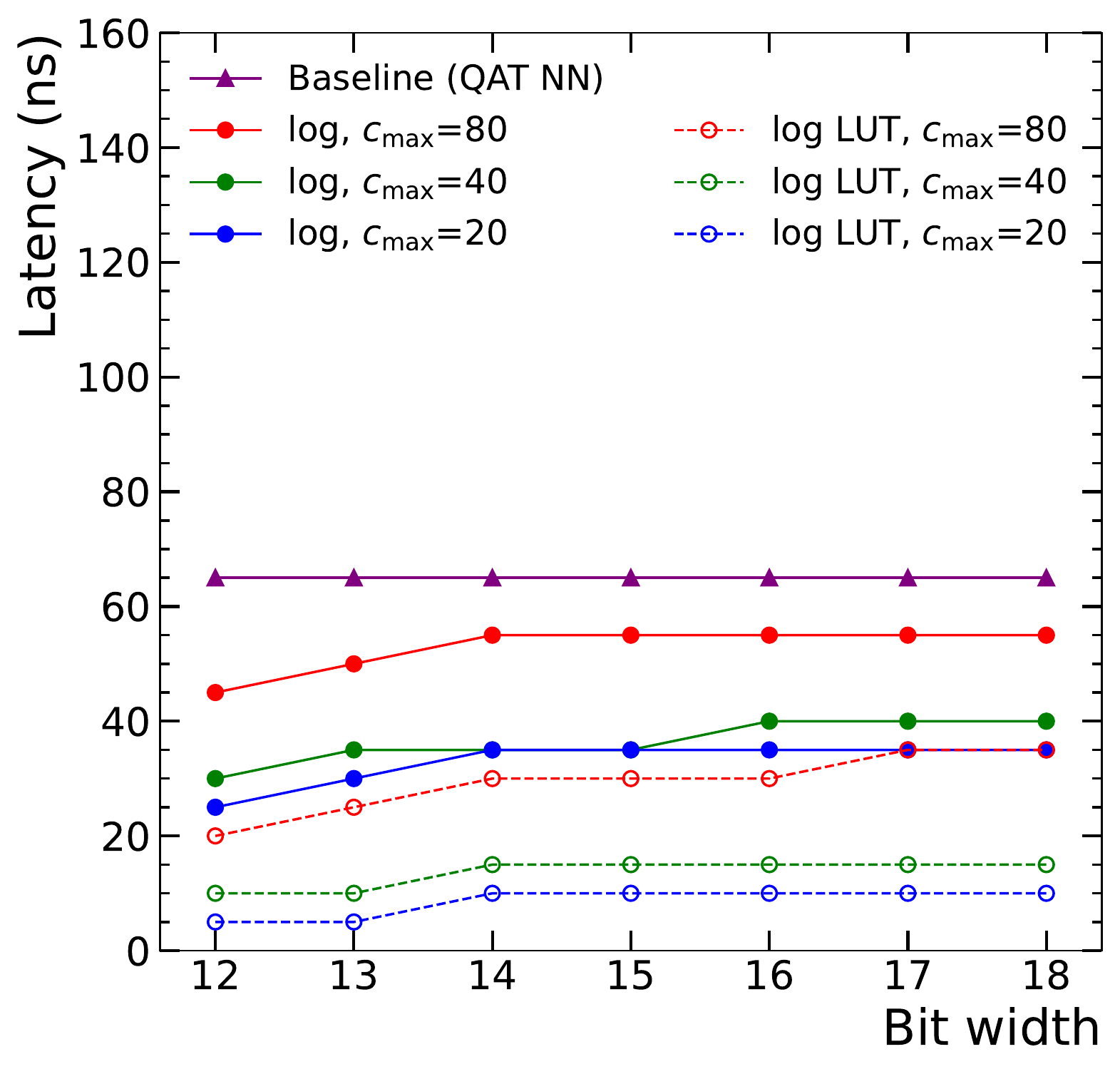}
    \caption{DSPs usage (left), LUTs usage (middle), and latency (right) as a function of bit width. From top to bottom: polynomial, trigonometric, exponential, and logarithmic models. The baseline QAT NN trained and implemented at corresponding precision is shown for comparison. Resource usage and latency are obtained from C-synthesis on a Xilinx VU9P FPGA with part number `xcvu9p-flga2577-2-e'.}
    \label{fig:plainandLUT_resource}
  \end{center}
\end{figure}

\begin{figure}[!h]
  \begin{center}
    \includegraphics[width=0.6\textwidth]{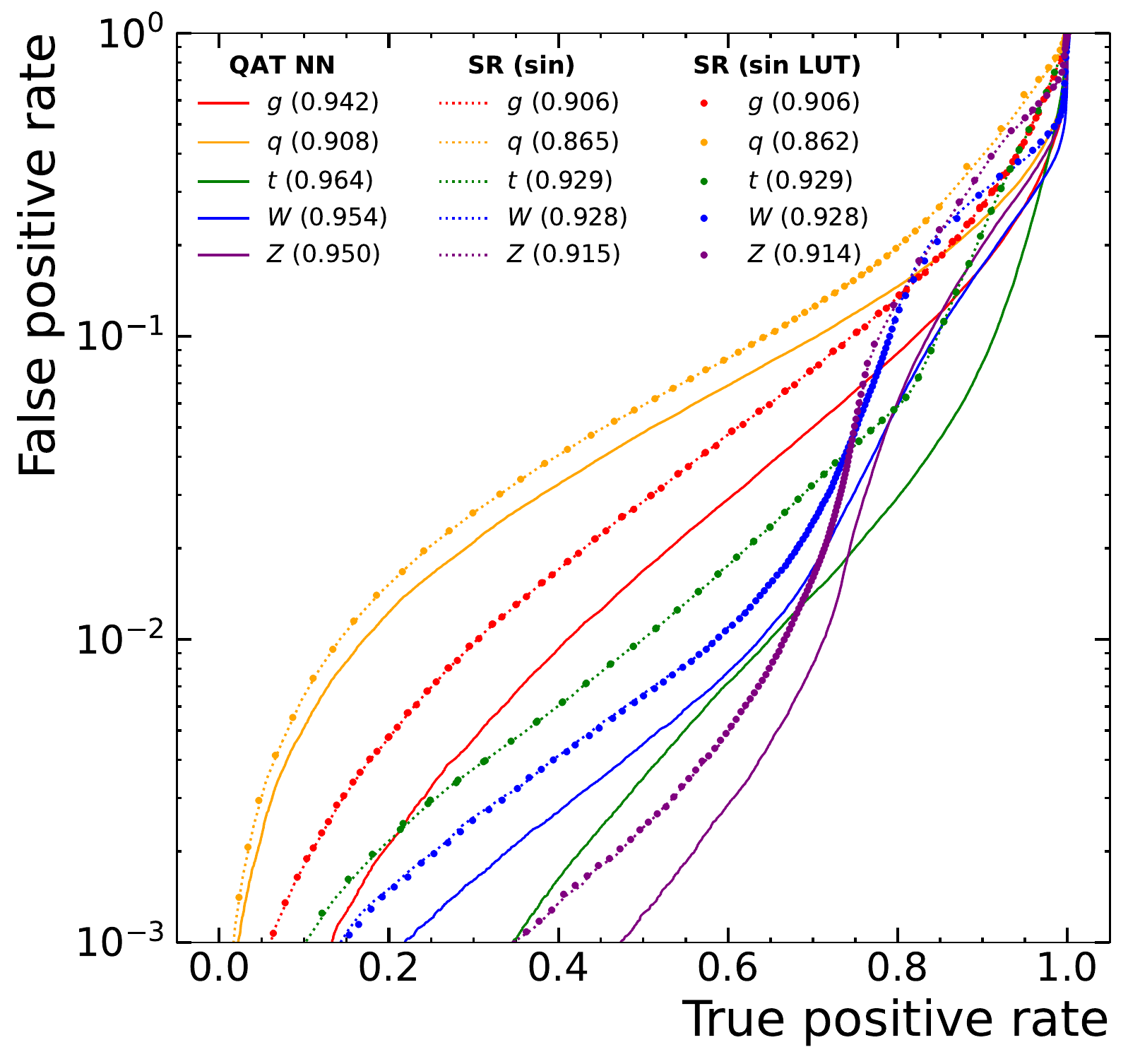}
    \caption{ROC curves for the trigonometric models with $c_{\text{max}}=80$ implemented with precision $\langle\text{16},\text{6}\rangle$, as compared to the baseline QAT NN. Numbers in parentheses correspond to the AUC per class.}
    \label{fig:sine_roc}
  \end{center}
\end{figure}

Based on the models in Section~\ref{SEC:plain} (except for the polynomial), we approximate all mathematical functions with LUTs and perform the analysis again.
In Fig.~\ref{fig:plainandLUT_acc} and~\ref{fig:plainandLUT_resource}, these models correspond to the dashed lines.
Compared to the baseline, the resource usage is dramatically reduced for all SR models, especially for those applying function approximation, sometimes with several orders of magnitude improvements.
Besides, SR models require significantly shorter inference time than the baseline, while having minimal drop in accuracy.
In particular, the inference time is reduced to as low as 1 clock cycle (5 ns) in certain scenarios in the exponential and logarithmic models with LUT-based functions implemented, which amounts to a reduction by a factor of 13 compared to the baseline, which has a latency of 13 clock cycles (65 ns), while still maintaining a relative accuracy above 90\%.
The ROC curves of the baseline and trigonometric models are compared in Fig.~\ref{fig:sine_roc}.

\subsection{Latency-aware training}

Alternatively, resource usage can be improved by guiding \pysr to search in a latency-aware manner.
By default, \pysr assigns the complexity for every operator to 1 so that they are all equally penalized when added to expression trees.
However, it is not ideal for FPGA deployment since, for example, an operator {\tt$\text{tan}(\cdot)$} typically takes several times more clock cycles than an {\tt$\text{sin}(\cdot)$} to evaluate on an FPGA.
This time cost can be incorporated in expression searches by setting operator complexity to the corresponding number of clock cycles needed on FPGAs.
Note that this strategy is not valid in the context of function approximation with LUTs since every indexing operation requires only one clock cycle.

\begin{figure}[!h]
  \begin{center}
    \includegraphics[width=0.329\textwidth]{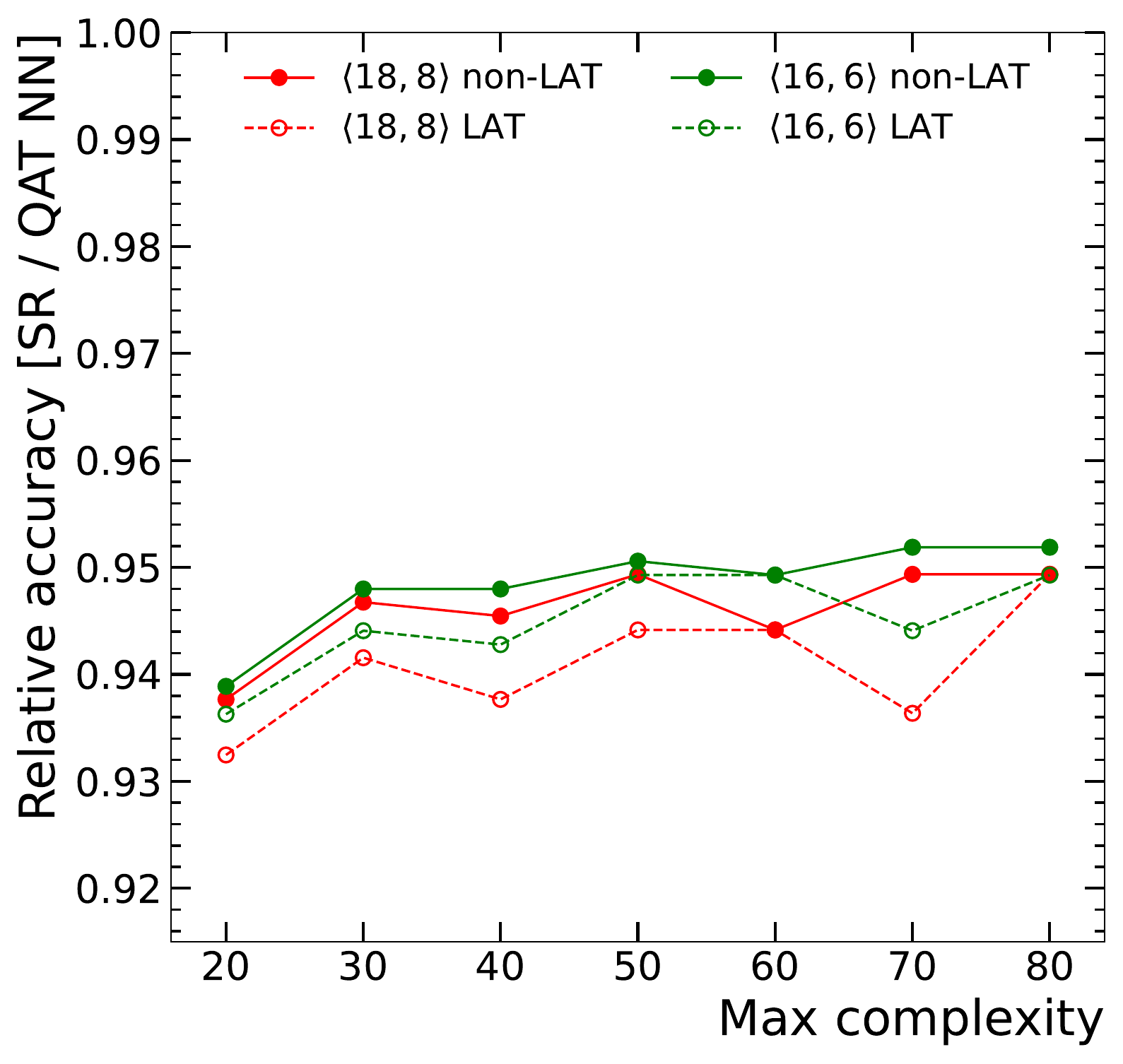}\\
    \includegraphics[width=0.329\textwidth]{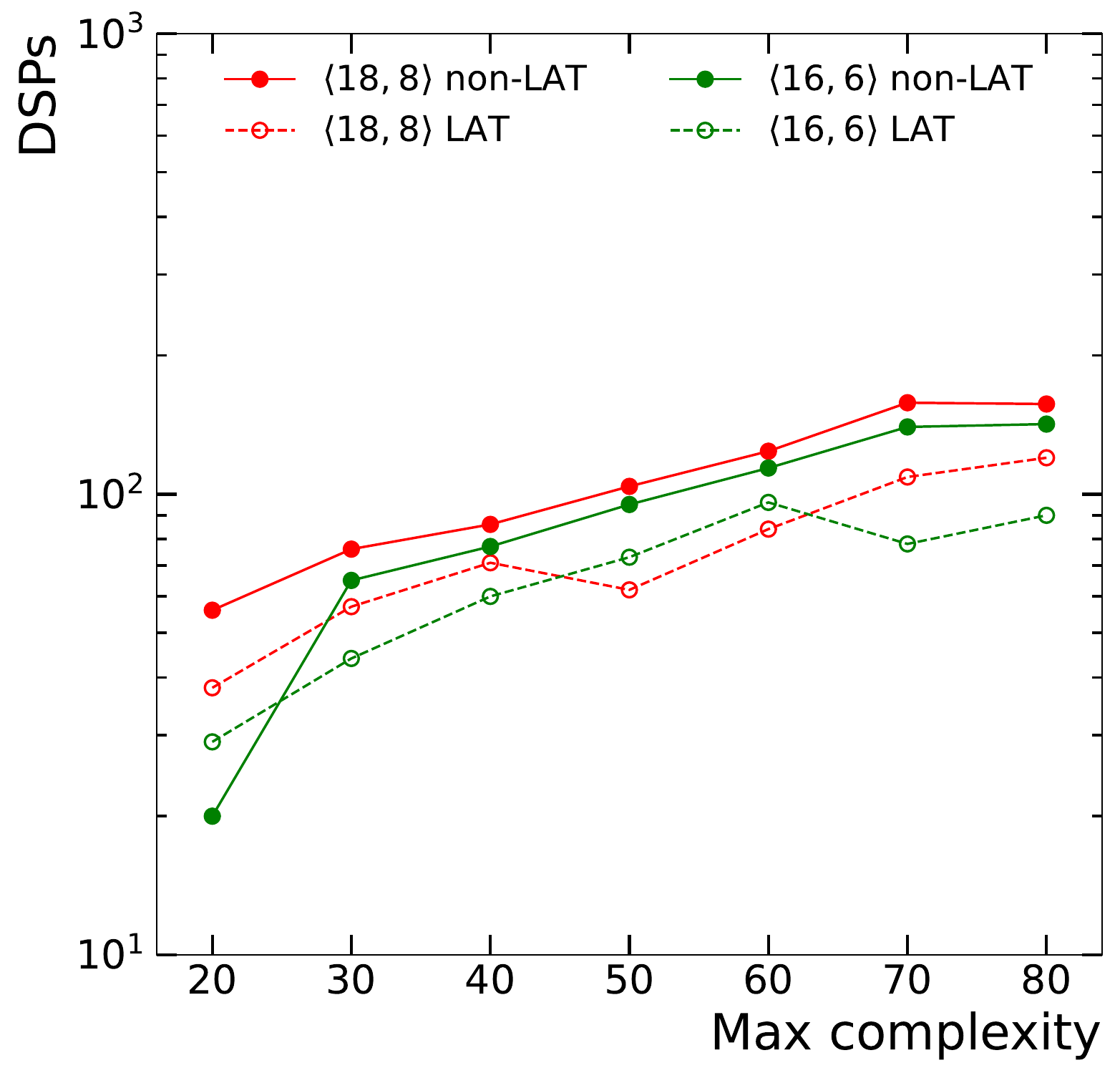}
    \includegraphics[width=0.329\textwidth]{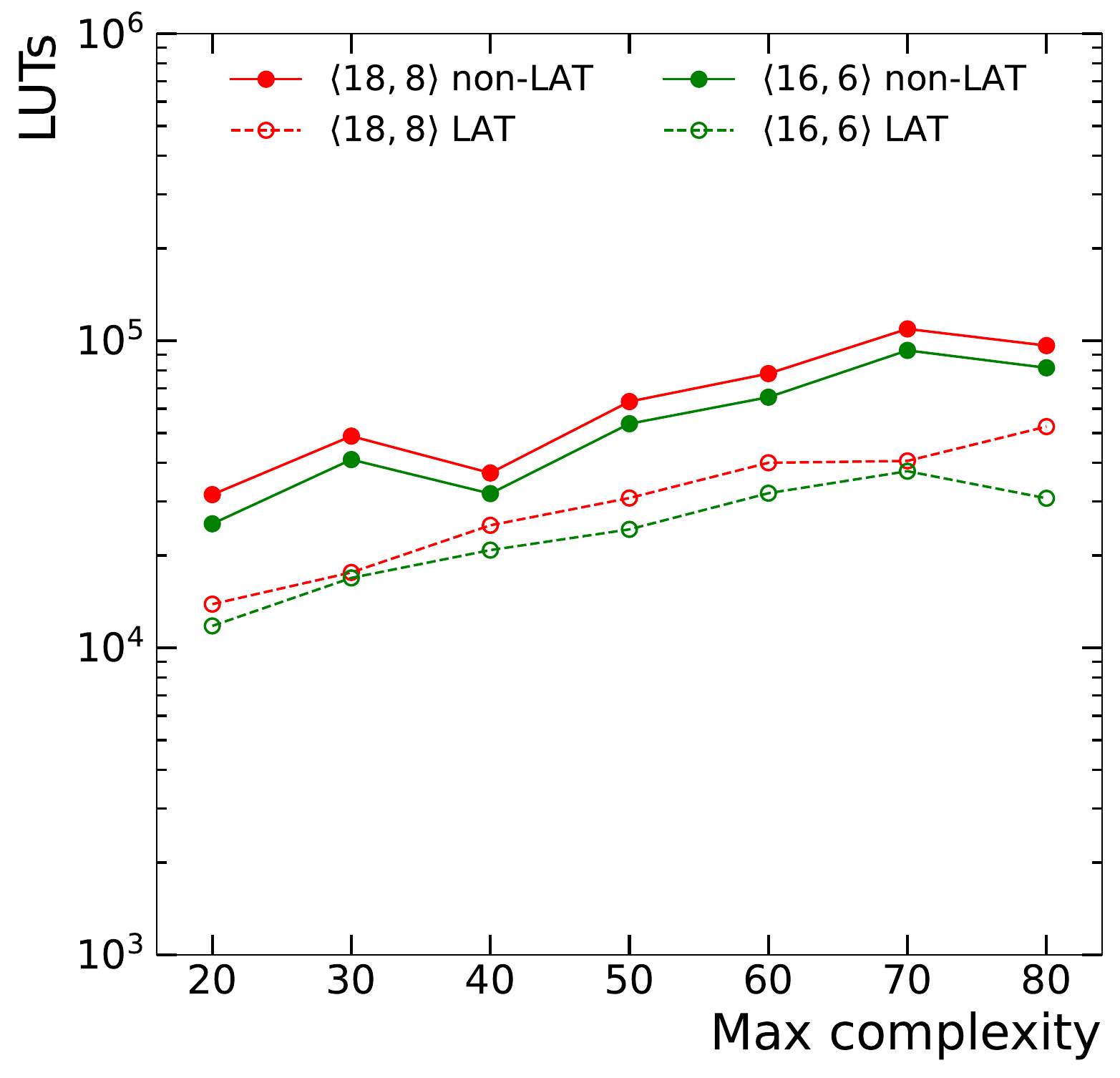}
    \includegraphics[width=0.329\textwidth]{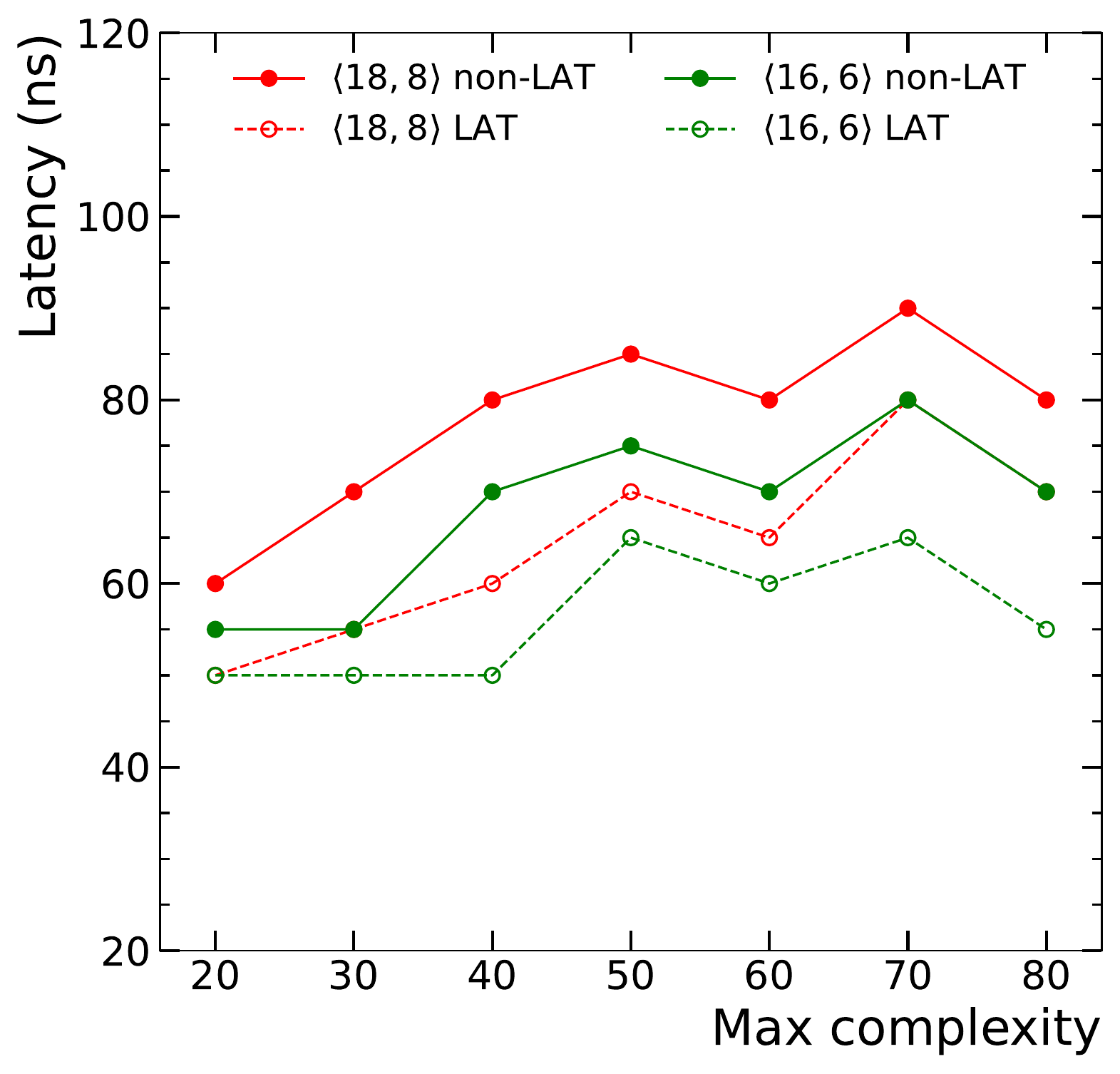}
    \caption{Relative accuracy (top), DSPs usage (bottom left), LUTs usage (bottom middle) and latency (bottom right) as a function of $c_{\text{max}}$ ranging from 20 to 80, comparing models obtained from plain implementation (solid) and LAT (dashed). Two precision settings are implemented: $\langle\text{16},\text{6}\rangle$ and $\langle\text{18},\text{8}\rangle$. The relative accuracy is evaluated with respect to the baseline model. Resource usage and latency are obtained from C-synthesis on a Xilinx VU9P FPGA with part number `xcvu9p-flga2577-2-e'.}
    \label{fig:LAT}
  \end{center}
\end{figure}

We demonstrate this latency-aware training (LAT) for two precisions, $\langle\text{16},\text{6}\rangle$ and $\langle\text{18},\text{8}\rangle$, with $c_{\text{max}}$ ranging from 20 to 80.
We consider the following operators: 
{\tt +}(1), {\tt -}(1), {\tt $\times$}(1), {\tt$\text{log}(\text{abs}(\cdot))$}(4), {\tt$\text{sin}(\cdot)$(}8), {\tt$\text{tan}(\cdot)$}(48), {\tt$\text{cosh}(\cdot)$}(8), {\tt$\text{sinh}(\cdot)$}(9), and {\tt$\text{exp}(\cdot)$}(3), where the numbers in parentheses correspond to the operator complexity for $\langle\text{16},\text{6}\rangle$ as an example.
For simplicity, function nesting is not allowed again. We also constrain the total complexity of the subtrees. In this way, we force the model to explore solutions in a different part of the Pareto front.
The final expressions are shown in Table~\ref{tab:expr_LAT}. Model accuracy, resource usage, and latency are shown in Fig.~\ref{fig:LAT}.
SR models obtained from LAT use systematically fewer resources and have a lower latency compared to those obtained from plain implementation, while having comparable accuracy. Implementation of a maximum-latency constraint is also possible. We added a script to generate operator complexity for praticioners\footnote{\url{https://github.com/AdrianAlan/hls4sr-configs}}.

\begin{table}[h!]
  \centering
  \resizebox{\textwidth}{!}{
    \begin{tabular}{ L | L | L }
      \hhline{===}
      \text{Operator complexity} & \text{Expression for the $t$ tagger with $c_{\text{max}}=40$} & \text{AUC} \\ \hhline{===}
      \text{All\ 1's\ (\pysr default)} & 0.11(C_{1}^{\beta=1} + C_{1}^{\beta=2} + \text{log}(\text{abs}(C_{1}^{\beta=2}))) - 0.48m_{\text{mMDT}} - 0.05\text{Multiplicity}(\text{Multiplicity} + \text{log}(\text{abs}(m_{\text{mMDT}}))) & 0.930 \\ & - \text{sin}(-C_{1}^{\beta=2} + 0.14C_{2}^{\beta=1}m_{\text{mMDT}}) + 0.11\text{sinh}(C_{1}^{\beta=1}) - 0.24 & \\ \hline
      \text{No.\ of\ clock\ cycles} & 0.04((\sum z\,\text{log}\,z) + C_{1}^{\beta=1} + C_{2}^{\beta=1} - m_{\text{mMDT}} - (\text{Multiplicity} - 0.2)(\text{Multiplicity} + \text{log}(\text{abs}(C_{1}^{\beta=2})))) & 0.924\\ \text{at }\langle16,6\rangle & - \text{sin}(- C_{1}^{\beta=1} - C_{1}^{\beta=2} + 1.23m_{\text{mMDT}} + 0.58) & \\\hline
      \text{No.\ of\ clock\ cycles} & 0.04\text{Multiplicity}(C_{1}^{\beta=2}(C_{1}^{\beta=2}-m_{\text{mMDT}}) - \text{Multiplicity} - \text{log}(\text{abs}(C_{1}^{\beta=2}((\sum z\,\text{log}\,z) + 0.23)))) & 0.926 \\ \text{at }\langle18,8\rangle & - \text{sin}(- C_{1}^{\beta=1} - C_{1}^{\beta=2} + 1.19m_{\text{mMDT}} + 0.61) & \\
      \hhline{===}
    \end{tabular}
  }
  \caption{Expressions generated by \pysr for the $t$ tagger with $c_{\text{max}}=40$, implemented with and without LAT. Constants are rounded to two decimal places for readability.}
  \label{tab:expr_LAT}
\end{table}

\clearpage

\section{Summary and Outlook}\label{SEC:Summary}

In this paper, we have presented a novel end-to-end procedure to utilize symbolic regression (SR) in the context of FPGAs for fast machine learning inference.
We extended the functionality of the \hlsfml package to support the expressions generated by \pysr.
We demonstrated the effectiveness of our approach on a physics benchmark (jet tagging at the LHC) and showed that our implementation of SR on FPGAs provides a way to dramatically reduce the computational resources needed to perform critical tasks, making it a promising alternative to deep learning models.
The utilization of SR in HEP provides a solution to meet the sensitivity and latency demands of modern physics experiments.
The results of this study open up new avenues for future work, including further optimization of the performance-resource trade-off and the exploration of other application domains for SR on FPGAs.

\section{Acknowledgments}\label{SEC:Acknowledgments}
We acknowledge the Fast Machine Learning collective as an open community of multi-domain experts and collaborators.
This community was important to the development of this project.
H.F.T. and S.D. are supported by the U.S. Department of Energy (Award No. DE-SC0017647). A.A.P. is supported by the Eric and Wendy Schmidt Transformative Technology Fund.
V.L. and P.H. are supported by A3D3 (NSF 2117997).
P.H. is also supported by the IAIFI grant.
M.P. is supported by the European Research Council (ERC) under the European Union’s Horizon 2020 research and innovation program (Grant No. 772369). 

\bibliography{reference}

\end{document}